\pdfoutput=1

\documentclass{article}
\usepackage{ijcai25}

\usepackage{times}
\usepackage{soul}
\usepackage{url}
\usepackage[hidelinks]{hyperref}
\usepackage[utf8]{inputenc}
\usepackage[small]{caption}
\usepackage{graphicx}
\usepackage{amsmath}
\usepackage{amsthm}
\usepackage{booktabs}
\usepackage{algorithm}
\usepackage{algorithmic}
\usepackage[switch]{lineno}

\usepackage{amssymb}
\usepackage{xcolor}
\definecolor{lightlightgray}{RGB}{217,217,217}
\usepackage{paralist}
\usepackage{subcaption}
\usepackage{multirow}
\usepackage[normalem]{ulem}
\usepackage{amsfonts}
\useunder{\uline}{\ul}{}
\usepackage{colortbl}
\usepackage[linesnumbered,ruled,algo2e]{algorithm2e}

\usepackage{pythonhighlight}

\newcommand{\hh}[1]{{\small\color{red}{\bf hh: #1}}}
\newcommand{\zhe}[1]{{\small\color{orange}{\bf zhe: #1}}}
\newcommand{\yz}[1]{{\small\color{blue}{\bf yz: #1}}}
\newcommand{\mh}[1]{{\small\color{cyan}{\bf mh: #1}}}
\newcommand{\hy}[1]{{\small\color{brown}{\bf hy: #1}}}
\newcommand{\yc}[1]{{\small\color{green}{\bf yuchen: #1}}}
\newcommand{\cleancomments}{
	\renewcommand{\hh}[1]{}
	\renewcommand{\zhe}[1]{}
	\renewcommand{\yz}[1]{}
        \renewcommand{\mh}[1]{}
        \renewcommand{\hy}[1]{}
        \renewcommand{\yc}[1]{}
}

\cleancomments

\title{Fine-grained Graph Rationalization}

\author{
Zhe Xu$^1$\and
Menghai Pan$^2$\and
Yuzhong Chen$^2$\and
Huiyuan Chen$^2$\and\\
Yuchen Yan$^1$\and
Mahashweta Das$^2$\and
Hanghang Tong$^1$\\
\affiliations
$^1$University of Illinois Urbana-Champaign\\
$^2$Visa Research\\
\emails
Corresponding: zhexu3@illinois.edu
}
\begin{document}

\maketitle

\begin{abstract}
Rationale discovery is defined as finding a subset of the input data that maximally supports the prediction of downstream tasks. In the context of graph machine learning, graph rationale is defined as locating the critical subgraph in the given graph topology. In contrast to the rationale subgraph, the remaining subgraph is named the environment subgraph. Graph rationalization can enhance the model performance as the mapping between the graph rationale and prediction label is viewed as invariant, by assumption. To ensure the discriminative power of the extracted rationale subgraphs, a key technique named \emph{intervention} is applied whose heart is that given changing environment subgraphs, the semantics from the rationale subgraph is invariant, which guarantees the correct prediction result. However, most, if not all, of the existing graph rationalization methods develop their intervention strategies on the graph level, which is coarse-grained. In this paper, we propose \underline{FI}ne-grained \underline{G}raph rationalization (FIG). Our idea is driven by the self-attention mechanism which provides rich interactions between input nodes. Based on that, FIG can achieve node-level and virtual node-level intervention. Our experiments involve $7$ real-world datasets, and the proposed FIG shows significant performance advantages compared to $13$ baseline methods.
\end{abstract}

% Web version:
% Rationale discovery is defined as finding a subset of the input data that maximally supports the prediction of downstream tasks. In the context of graph machine learning, graph rationale is defined to locate the critical subgraph in the given graph topology. In contrast to the rationale subgraph, the remaining subgraph is named the environment subgraph. Graph rationalization can enhance the model performance as the mapping between the graph rationale and prediction label is viewed as invariant, by assumption. To ensure the discriminative power of the extracted rationale subgraphs, a key technique named "intervention" is applied whose heart is that given changing environment subgraphs, the semantics from the rationale subgraph is invariant, which guarantees the correct prediction result. However, most, if not all, of the existing graph rationalization methods develop their intervention strategies on the graph level, which is coarse-grained. In this paper, we propose fine-grained graph rationalization (FIG). Our idea is driven by the self-attention mechanism which provides rich interactions between input nodes. Based on that, FIG can achieve node-level and virtual node-level intervention. Our experiments involve 7 real-world datasets, and the proposed FIG shows significant performance advantages compared to 13 baseline methods.

\section{Introduction}

% \hh{see if this paper is relevant \url{https://arxiv.org/abs/2310.19035}}
% \zhe{Thanks, Dr. Tong. I will check it soon.}
Rationale refers to a subset of the input features that play a crucial role in model predictions for downstream tasks~\cite{DBLP:conf/icml/ChangZYJ20,DBLP:conf/kdd/LiuZXL022,DBLP:conf/iclr/WuWZ0C22,DBLP:conf/cikm/ZhangDT22,DBLP:conf/www/YueLLGYL24,DBLP:conf/nips/SuiWWCLZW023}. In the context of graph machine learning, graph rationale is defined as a subgraph of the input graph containing the most task-relevant semantics. 

% Additionally, the discovery of rationales proves to be beneficial in enabling effective predictions on out-of-distribution test samples.
%   This benefit can be attributed to the common content/style decomposition~\cite{DBLP:journals/corr/abs-2206-15475}, which reveals that data from different distributions may exhibit changing style (i.e., environment) features, while their content features (i.e., rationales) remain invariant.

The application of graph rationale is broad, for example, it can greatly enhance model performance for graph-level tasks~\cite{DBLP:conf/iclr/WuWZ0C22} by identifying the key components of the input graph. Additionally, the discovery of rationales can improve model explainability~\cite{DBLP:conf/kdd/LiuZXL022}, as it highlights the parts of the input graph that significantly contribute to the final prediction.

% \begin{figure}[t!]
% 	\centering
% 	\includegraphics[width=0.47\textwidth]{Figures/illustration.png}
% 	\caption{Illustration of the rationalization and intervention. The intervention is the key to ensure the rationale $i$ truly has the discriminative power for the label $i$.}
% 	\label{fig: rationalization illustration}
% \end{figure}

\begin{figure}[t!]
	\centering
	\includegraphics[width=\columnwidth]{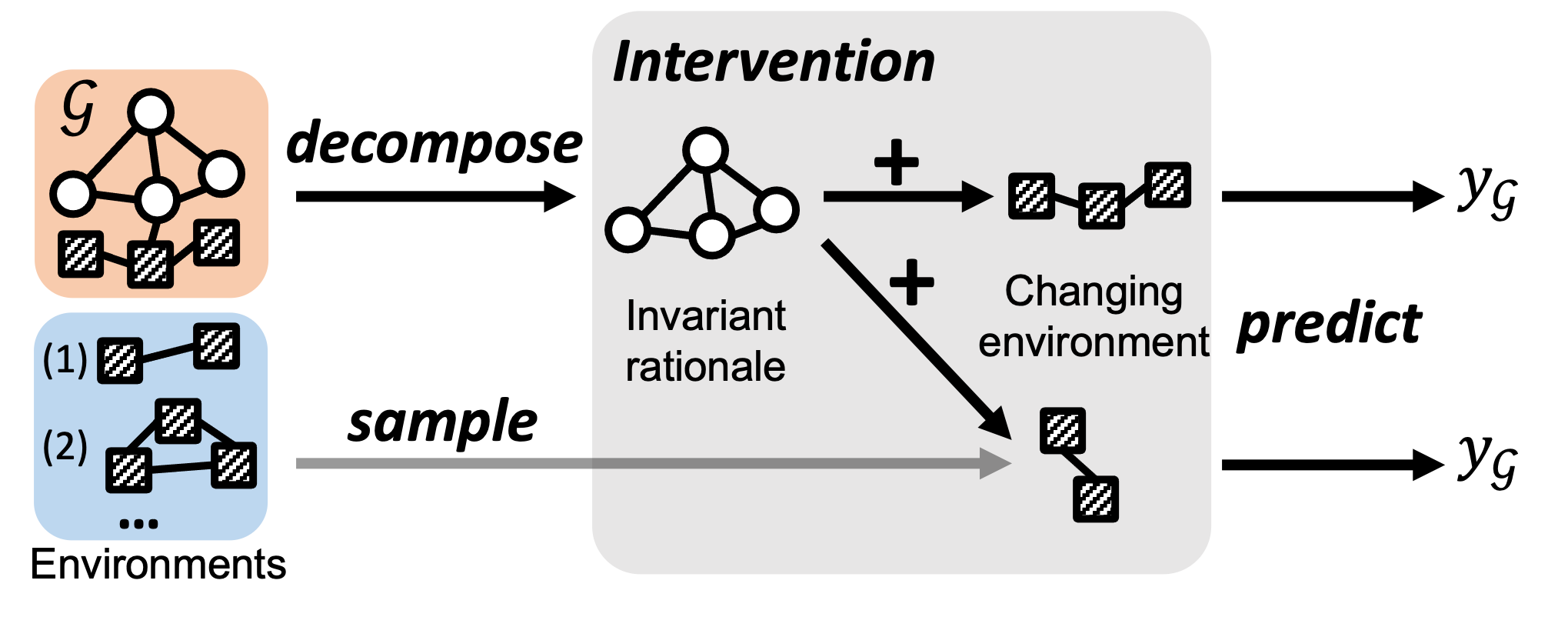}
	\caption{Illustration of the rationale/environment decomposition and intervention. Round nodes denote graph rationales, and square nodes (with stripes) denote the environments. The intervention aims to ensure the rationale from graph $\mathcal{G}$ truly has the discriminative power for the label $y_{\mathcal{G}}$.}
	% \Description{TODO.}
	\label{fig: rationalization illustration2}
\end{figure}

Existing graph rationalization solutions~\cite{DBLP:conf/kdd/LiuZXL022,DBLP:conf/iclr/WuWZ0C22} employ a trainable \emph{augmenter} to execute the rationale/environment decomposition. In this process, a node/edge mask is generated by the augmenter to decompose the given graph into a rationale graph and an environment graph. Inspired by the content-style decomposition~\cite{DBLP:journals/corr/abs-2206-15475}, the key idea of graph rationalization is to preserve the utility of the graph rationale even when faced with changing environment graphs (see Figure~\ref{fig: rationalization illustration2}). To achieve this, a technique named \emph{intervention} is used, where the environment graph interacts with the rationale graph.

The intervention mechanism (named \emph{intervener}) is essential in the graph rationalization process, as it must accurately represent the interaction between the rationale and the environment. Intuitively, the intervener should work in an adversarial behavior against the aforementioned augmenter, a point not emphasized in the existing literature. If the intervener is more powerful, it can capture more detailed interactions between the rationale and environment subgraphs. Given such a powerful intervener, the augmenter is compelled to minimize these interactions between the graph rationale and the environment to obtain a ``purer'' graph rationale.

Unfortunately, existing works develop interveners in a coarse and non-parametric manner. After performing rationale/environment decomposition on the graph data, they compute graph-level embeddings for the rationale and environment subgraphs. The intervention is then designed as an interaction between these graph-level embeddings. For example,~\cite{DBLP:conf/kdd/LiuZXL022} adds the environment embedding into the rationale embedding as the intervened rationale embedding; ~\cite{DBLP:conf/iclr/WuWZ0C22} defines the intervened prediction as the Hadamard product between the predictions based on the rationale subgraph and the environment subgraph. We argue that such a graph-level non-parametric intervention is insufficient to represent the interaction between the rationale and environment graphs effectively.

In response to this limitation, we propose a fine-grained, parametric intervention mechanism named \underline{FI}ne-grained \underline{G}raph rationalization (FIG). Our proposed FIG draws inspiration from the self-attention module in the Transformer model, which captures interactions between input tokens. Building upon insights from Transformer~\cite{DBLP:conf/nips/VaswaniSPUJGKP17} and its linear variant Linformer~\cite{DBLP:journals/corr/abs-2006-04768}, FIG formulates the interaction between the rationale and environment subgraphs at the node-level or the virtual node-level. The two variants are named FIG-N and FIG-VN. Additionally, to maximize the effectiveness of the intervention, we formulate a min-max game involving the node encoder, augmenter, intervener, and predictor, compelling the rationale subgraph to be as informative as possible.

We conduct comprehensive experiments on $7$ graph-level benchmarks to evaluate the proposed approach and compare FIG-N/VN against $13$ state-of-the-art baseline methods. The results demonstrate that our proposed FIG and its variants outperform the baseline methods, validating their superior performance. Our primary contributions in this paper are summarized as follows:
\begin{itemize}
    \item We address the graph rationalization problem via a fine-grained model FIG, which works at the node/virtual node level.
    \item A min-max objective function is proposed so that the effectiveness of the proposed intervener can be maximized.
    \item Extensive experiments covering $13$ baseline methods and $7$ real-world datasets are conducted.
    % to verify the proposed FIG's efficacy.
\end{itemize}

\section{Preliminaries}
\label{sec: preliminaries}

This section first introduces the notations used throughout this paper~\ref{sec: notations}. Then, the classic Transformer architecture is introduced in Section~\ref{sec: graph transformer}, whose self-attention module is the important building block of our model. Last but not least, we introduce the overall ideas of existing graph rationalization works in Section~\ref{sec: invariant rationalization preliminary}.
\subsection{Notations}
\label{sec: notations}
We adopt the following notation conventions: bold uppercase letters for matrices and tensors (e.g., $\mathbf{A}$), bold lowercase letters for column vectors (e.g., $\mathbf{u}$), lowercase and uppercase letters in regular font for scalars (e.g., $d$, $K$), and calligraphic letters for sets (e.g., $\mathcal{T}$). To index vectors/matrices/tensors, we follow the syntax from NumPy\footnote{\url{https://numpy.org/doc/stable/index.html}} ($0$-based). Specifically, $\mathbf{A}[p,:]$ and $\mathbf{A}[:,q]$ represent the $p$-th row and the $q$-th column of matrix $\mathbf{A}$ respectively; $\mathbf{A}[p,q]$ represents the entry at the $p$-th row and the $q$-th column. Similarly, $\mathbf{u}[p]$ denotes the $p$-th entry of vector $\mathbf{u}$. In addition, the slicing syntax for vectors/matrices/tensors is used. For example, for a matrix $\mathbf{A}$, $\mathbf{A}[i:j,:]$ denotes rows from the $i$-th row (included) to the $j$-th row (excluded) and $\mathbf{A}[:,:k]$ denotes all the columns before the $k$-th column. The superscript $\top$ denotes the transpose of matrices and vectors. $\odot$ represents the Hadamard product, and $\circ$ denotes function composition. We use $||$ to represent the concatenation operation and the specific dimension of concatenation will be clarified based on the context.

An attributed graph can be represented as $\mathcal{G} = (\mathbf{A}, \mathbf{X}, \mathbf{E})$, where $\mathbf{A}\in\mathbb{R}^{n\times n}$ is the adjacency matrix, $\mathbf{X}\in\mathbb{R}^{n\times d_X}$ is the node feature matrix, and $\mathbf{E}\in\mathbb{R}^{n\times n\times d_E}$ is the edge feature tensor. Here, $n$ denotes the number of nodes, and $d_X$ (or $d_E$) represents the dimensions of node (or edge) features, respectively. This paper assumes the node and edge feature dimensions are the same (i.e., $d_X=d_E=d$) for brevity; if they differ, a simple, fully connected layer can map them into a common feature space. Our main focus in this paper is on graph property prediction tasks. The ground truth of a graph is represented by $y$.
% It is important to note that (1) either node features $\mathbf{X}$ or edge features $\mathbf{E}$ or both can be missing and here we present a graph as $\mathcal{G} = (\mathbf{A}, \mathbf{X}, \mathbf{E})$ for completeness and (2) 

\subsection{Graph Transformer}
\label{sec: graph transformer}

The core modules of the Transformer architecture~\cite{DBLP:conf/nips/VaswaniSPUJGKP17} are the self-attention layer and the feed-forward network layer. Given the input as a sequence of symbol representations $\mathbf{H}\in\mathbb{R}^{n\times d_H}$, it is first transformed into the query, key, and value matrices as
\begin{equation}
    \mathbf{Q} = \mathbf{H}\mathbf{W}_{Q}, \mathbf{K} = \mathbf{H}\mathbf{W}_{K}, \mathbf{V} = \mathbf{H}\mathbf{W}_{V},\label{eq: query, key, value computation}
\end{equation}
where $\mathbf{W}_{Q}\in\mathbb{R}^{d_H\times d_Q}$, $\mathbf{W}_{K}\in\mathbb{R}^{d_H\times d_K}$, $\mathbf{W}_{V}\in\mathbb{R}^{d_H\times d_V}$. For the brevity of the presentation, we set $d_H=d_Q=d_K=d_V=d$. Then, the self-attention module works as,
\begin{subequations}
\begin{align}
    \mathbf{P} &= \mathtt{Attn}(\mathbf{H}) = \sigma(\frac{\mathbf{Q}\mathbf{K}^{\top}}{\sqrt{d}}),\label{eq: attention computation}\\
    \mathbf{H}&\leftarrow \mathbf{P}\mathbf{V}+\mathbf{H}.\label{eq: residual in attention}
\end{align}
\end{subequations}
Typically, the non-linearity $\sigma$ is $\mathtt{Softmax}$. The feed-forward network (FFN) updates the symbol representations $\mathbf{H}$ as:
\begin{equation}
    \mathbf{H} \leftarrow \mathtt{FFN}(\mathbf{H})+\mathbf{H}\label{eq: feed-forward network}.
\end{equation}
Additional techniques such as layer/batch normalization~\cite{DBLP:journals/corr/BaKH16,DBLP:conf/icml/IoffeS15}, dropout~\cite{DBLP:journals/jmlr/SrivastavaHKSS14}, and multi-head attention~\cite{DBLP:conf/nips/VaswaniSPUJGKP17} can be included, but omitted here for brevity.

While Transformers was originally devised for sequence or set data with positional encoding, numerous techniques have since been introduced to adapt Transformers for graph data. Based on the taxonomy outlined by~\cite{DBLP:journals/corr/abs-2202-08455}, most graph Transformers are designed from the perspectives of (1) incorporating the topology encoding into the node features, (2) incorporating the topology encoding into the attention matrix, and (3) utilizing graph neural networks~\cite{DBLP:journals/tnn/WuPCLZY21} as auxiliary modules.

% (e.g., $\mathbf{H}=\mathbf{X}||\texttt{Encode}(\mathbf{A})$) where a typical example of $\texttt{Encode}$ function is extracting the eigenvectors, (2) incorporating the topology encoding into the attention matrix (e.g., $\mathbf{P}=\texttt{COMBINE}(\frac{\mathbf{Q}\mathbf{K}^{\top}}{\sqrt{d}},\mathbf{A})$) where an example of $\texttt{COMBINE}$ function is Hadamard product, and (3) utilizing graph neural networks (GNNs)~\cite{DBLP:journals/jmlr/DwivediJL0BB23} as an auxiliary module (e.g., $\mathbf{H}=\texttt{GNN}(\mathbf{A},\mathbf{X})$).

Interestingly, it is well-known in both the graph learning~\cite{DBLP:conf/icml/ChenOB22} and natural language processing communities~\cite{DBLP:conf/nips/ZaheerGDAAOPRWY20,DBLP:journals/corr/abs-2006-04768} that, from the message-passing perspective, the key idea of the Transformer architecture is to reconstruct a weighted complete graph, whose adjacency matrix is $\mathbf{P}=\sigma\left(\frac{\mathbf{Q}\mathbf{K}^{\top}}{\sqrt{d}}\right)$.

\begin{figure*}[t!]
\begin{subfigure}{\textwidth}
  \centering
  \includegraphics[width=0.68\textwidth]{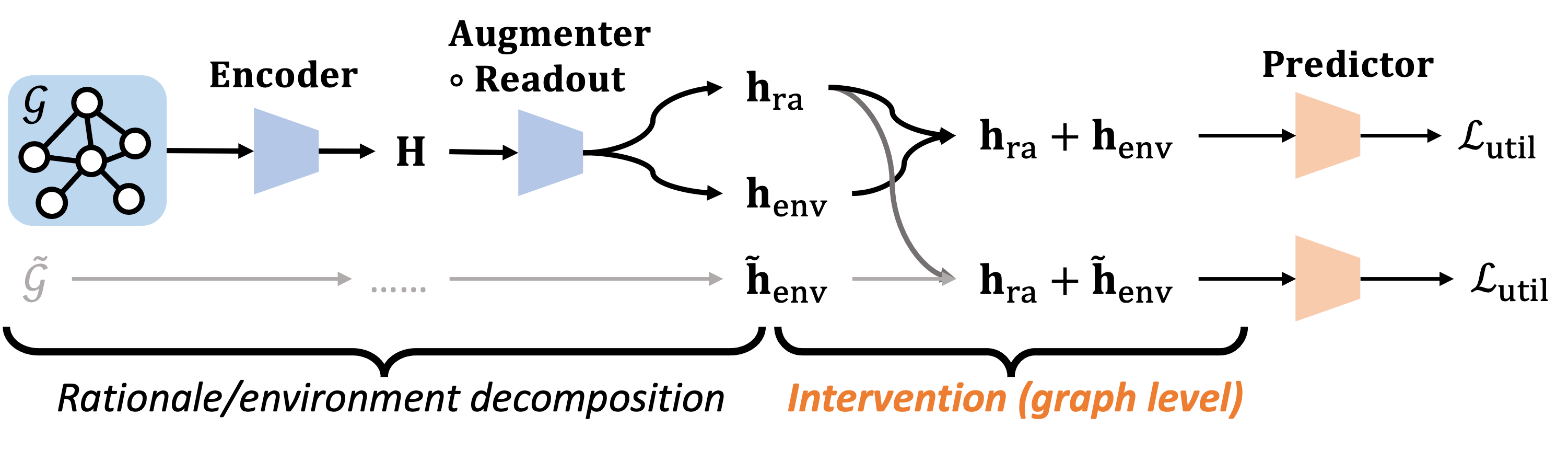}
  \caption{GREA~\cite{DBLP:conf/kdd/LiuZXL022}}
  \label{fig: GREA pipeline}
  % \Description{TODO.}
\end{subfigure}

\begin{subfigure}{\textwidth}

  \centering
  \includegraphics[width=0.8\textwidth]{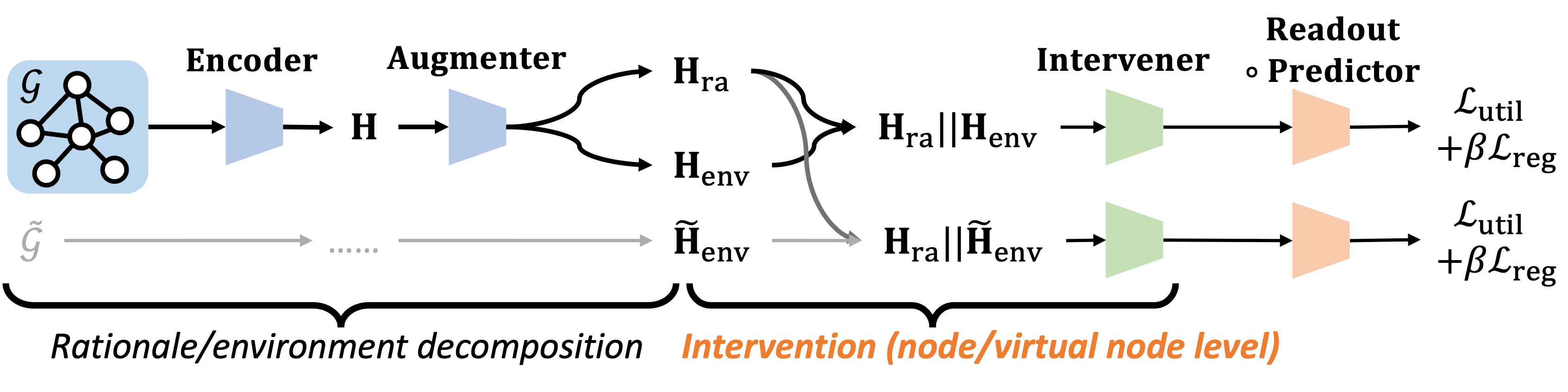}
  \caption{FIG (Ours)}
  \label{fig: FIG pipeline}
  % \Description{TODO.}
\end{subfigure}
\caption{Pipeline comparison between existing work GREA and proposed FIG. $\circ$ denotes function composition. GREA designs the intervention at the graph level, and the proposed FIG designs the intervention at the node/virtual node level. The augmented environment $\tilde{\mathbf{H}}_{\texttt{env}}$ is from another graph $\tilde{\mathcal{G}}$ (through the Encoder and Augmenter) in the batch.}
\end{figure*}

% \zhe{Check if we can learn something from GAN's theory.}

% \hh{(if we have time), draw an illustrative figure for G-tilde, similar to the one in figure 2. In fact, we might consider re-use (part of) figure 2 here: to mark which one is rationale, environment, etc.} \zhe{I prefer not, because the tilde G is only used for generating the environments (in practical implementations), its rationale part is totally useless in this overall figure (centered around the graph G). Let me know if you have other ideas.}

\subsection{Invariant Rationale Discovery on Graphs}
\label{sec: invariant rationalization preliminary}
The graph rationale is a subgraph that encodes most downstream task-relevant semantics. A typical example is the functional groups in polymer graphs~\cite{DBLP:conf/kdd/LiuZXL022,DBLP:conf/iclr/WuWZ0C22}, which fundamentally determines the chemical property of polymers. Mathematically, a given graph is decomposed into a rationale graph and an environment graph: $\mathcal{G}=\mathcal{G}_{\texttt{ra}}\cup\mathcal{G}_{\texttt{env}}$. Commonly, the graph embeddings on $\mathcal{G}_{\texttt{ra}}$ and $\mathcal{G}_{\texttt{env}}$ are computed as $\mathbf{h}_{\texttt{ra}}$ and $\mathbf{h}_{\texttt{env}}$. To ensure the rationale graph is invariant w.r.t. the prediction results when confronting different environments, a utility loss is minimized given the rationale embedding $\mathbf{h}_{\texttt{ra}}$ intervened by the environment embedding $\tilde{\mathbf{h}}_{\texttt{env}}$, i.e., $\min\mathcal{L}_{\texttt{util}}(\mathbf{h}_{\texttt{ra}}\xleftarrow{\text{intervene}}\tilde{\mathbf{h}}_{\texttt{env}})$. Here, $\tilde{\mathbf{h}}_{\texttt{env}}$ could be either from the same graph (i.e., $\tilde{\mathbf{h}}_{\texttt{env}}=\mathbf{h}_{\texttt{env}}$), or could be environment embeddings from other graphs, such as those in the batch. A key difference among existing methods lies in the intervention operation, of which we mention two: 
% (1) GREA~\cite{DBLP:conf/kdd/LiuZXL022} designs the intervention as adding operation, i.e., $\mathbf{h}_{\texttt{ra}}+\tilde{\mathbf{h}}_{\texttt{env}}$ and (2) DIR~\cite{DBLP:conf/iclr/WuWZ0C22} designs the intervention as an element-wisely weighting of prediction: $\theta_{\texttt{pred}}(\mathbf{h}_{\texttt{ra}})\odot\mathtt{Sigmoid}(\theta_{\texttt{pred}}(\tilde{\mathbf{h}}_{\texttt{env}}))$, where $\odot$ is the Hadamard product and $\theta_{\texttt{pred}}$ is a predictor

\begin{itemize}
    \item GREA~\cite{DBLP:conf/kdd/LiuZXL022} designs the intervention as adding operation, i.e., $\mathbf{h}_{\texttt{ra}}+\tilde{\mathbf{h}}_{\texttt{env}}$;
    \item DIR~\cite{DBLP:conf/iclr/WuWZ0C22} designs the intervention as an element-wisely weighting of prediction: $\theta_{\texttt{pred}}(\mathbf{h}_{\texttt{ra}})\odot\mathtt{Sigmoid}(\theta_{\texttt{pred}}(\tilde{\mathbf{h}}_{\texttt{env}}))$, where $\odot$ is the Hadamard product and $\theta_{\texttt{pred}}$ is a predictor.
\end{itemize}
The above intervention is conducted at the graph level because $\mathbf h_{\texttt{ra}}$ and $\tilde{\mathbf f}_{\texttt{env}}$ are graph embeddings. In Figure~\ref{fig: GREA pipeline}, an overview of the GREA~\cite{DBLP:conf/kdd/LiuZXL022} is presented. As a comparison, we aim to design the intervention at finer grains (e.g., node level) to handle the interaction between rationale and environment graphs, which will be detailed as follows.

% For example, GREA~\cite{DBLP:conf/kdd/LiuZXL022} minimizes the utility loss of the prediction given $\mathbf{h}_{\texttt{ra}}+\mathbf{h}_{\texttt{env}}$. DIR~\cite{DBLP:conf/iclr/WuWZ0C22} designs the intervention on the predicted graph labels on rationale and environment graphs ($\hat{\mathbf{y}}_{\texttt{ra}}$ and $\hat{\mathbf{y}}_{\texttt{env}}$). To be specific, the utility loss is minimized given the intervened prediction $\hat{\mathbf{y}}_{\texttt{ra}}\odot\texttt{Sigmoid}(\hat{\mathbf{y}}_{\texttt{env}})$ where $\odot$ is the Hadamard product. In Figure~\ref{fig: GREA pipeline} we present an overview of the GREA~\cite{DBLP:conf/kdd/LiuZXL022} for illustration. \hh{would it be helpful if we talk about what intervention does in these two methods, e.g., perturbing the embedding of rationale by that of the environment (adding them together) in grea; element-wisely weighting the prediction vector of the rationale by that of the enviroment in dir?}

\section{Proposed Model}
\label{sec: method}

In this section, we introduce our proposed graph rationalization method, FIG. At its core, FIG utilizes a module based on the Transformer architecture. Figure~\ref{fig: FIG pipeline} provides an overview of FIG, highlighting its four main parametric modules: the encoder, augmenter, intervener, and predictor.

\noindent\textbf{Encoder.} The encoder, denoted as $\theta_{\texttt{enc}}:\mathcal{G}\rightarrow \mathbb{R}^{n\times d}$, accepts a graph data as input and produces a node embedding matrix as output. While there are various graph encoders available, such as graph neural networks (GNNs)~\cite{DBLP:journals/tnn/WuPCLZY21} and graph Transformers~\cite{DBLP:journals/corr/abs-2202-08455}. From the methodology perspective, the encoder module is not the main contribution of this paper, so in this section, we do not specify a specific graph encoder $\theta_{\texttt{enc}}$.

\noindent\textbf{Predictor.} The predictor, denoted as $\theta_{\texttt{pred}}:\mathbb{R}^{d}\rightarrow\mathbb{R}^c$ takes as input a graph embedding and outputs a task-related vector/scalar. For graph regression tasks, $c=1$; for graph classification tasks, $c$ is the number of classes. A typical choice of predictor is a multi-layer perceptron ($\mathtt{MLP}$) with appropriate activation functions. Details of the encoder and predictor in our implementation are presented in Section~\ref{sec: experiments}.

In subsequent subsections, we will elaborate on the augmenter and intervener, two essential modules. Their detailed designs derive two variants of the proposed FIG model.

% \zhe{Try to merge two methods into a uniform one.}
\subsection{Node-Level Variant: FIG-N}
\label{sec: FIG-N}

\noindent\textbf{Node-level augmenter.} The augmenter is a critical module of the proposed FIG. For the node-level variant, termed FIG-N, the augmenter's primary function is decomposing the node set into two distinct subsets: rationale nodes and environment nodes. This decomposition is operated by parameterizing the node-level augmenter as a learnable node partitioner, denoted by $\theta_{\texttt{aug-N}}$,
\begin{equation}
    \mathbf{m} = \mathtt{Sigmoid}(\mathtt{MLP}(\mathbf{H}, \theta_{\texttt{aug-N}})),\label{eq: node-level augmenter}
\end{equation}

\noindent whose input is the node embedding matrix $\mathbf{H}\in\mathbb{R}^{n\times d}$, and its output is a partition vector $\mathbf{m}\in[0,1]^{n}$. $\mathtt{MLP}$ is a multi-layer perceptron. Each entry within $\mathbf{m}$, such as $\mathbf{m}[i]$, signifies the probability of the $i$-th node being categorized as a rationale node.

For the node partition vector $\mathbf{m}$, its top-$K$ entries are indexed as $\texttt{idx}_{\texttt{ra}} = \mathtt{argtopK}(\mathbf{m})$ which is used to index the rationale nodes from the node embedding matrix $\mathbf{H}$; naturally, the remaining nodes are categorized as the environment nodes whose indices are $\texttt{idx}_{\texttt{env}}=\{1,\dots, n\}-\texttt{idx}_{\texttt{ra}}$. $K$ is a hyper-parameter whose impact is studied in Section~\ref{sec: sensitivity study}. Also, in our implementation, we use a soft $\mathtt{argtopK}$ operation to remain differentiability whose details are in Section~\ref{sec: argtopK}.

Using the indices mentioned above, rationale and environment embeddings, denoted as $\mathbf{H}_{\texttt{ra}}$ and $\mathbf{H}_{\texttt{env}}$, respectively, can be extracted from the node embedding matrix $\mathbf{H}$:
\begin{subequations}
\begin{align}
\mathbf{H}_{\texttt{ra}}&=\mathbf{H}[\texttt{idx}_\texttt{ra}, :]\in\mathbb{R}^{K\times d},\label{eq: rationale node embeddings}\\
\mathbf{H}_{\texttt{env}}&=\mathbf{H}[\texttt{idx}_\texttt{env}, :]\in\mathbb{R}^{(n-K)\times d},\label{eq: environment node embeddings}
\end{align}
\end{subequations}

\noindent\textbf{Node-level intervener.} The design of the fine-grained intervener draws inspiration from the Transformer architecture~\cite{DBLP:conf/nips/VaswaniSPUJGKP17}. Explicitly, the node-level intervener $\phi$ is presented as,
\begin{subequations}
\begin{align}
    \mathbf{H}_{\texttt{inter}}, \mathbf{P} &= \mathtt{Transformer}(\mathbf{H}_{\texttt{ra}}||\mathbf{H}_{\texttt{env}})\label{eq: intervener part 1},\\
    where\quad \mathbf{P} &= \mathtt{Attn}(\mathbf{H}_{\texttt{ra}}||\mathbf{H}_{\texttt{env}})\label{eq: intervener part 2}.
\end{align}
\end{subequations}
In this representation, the operator $||$ concatenates along the first dimension of the matrices $\mathbf{H}_{\texttt{ra}}$ and $\mathbf{H}_{\texttt{env}}$. We dub the Eqs.~\eqref{eq: query, key, value computation}-\eqref{eq: feed-forward network} as $\mathtt{Transformer}$ and $\mathbf{P}$ is the intermediate attention matrix from the self-attention layer (Eq.~\eqref{eq: intervener part 2}). Here, the self-attention module models the interactions between the rational nodes $\mathbf{H}_{\texttt{ra}}$ and the environment nodes $\mathbf{H}_{\texttt{env}}$. $\phi$ includes all the parameters of the $\mathtt{Attn}$ (Eq.~\eqref{eq: intervener part 2}) and $\mathtt{FFN}$ (Eq.~\eqref{eq: feed-forward network}) modules. In some contexts where the attention matrix $\mathbf{P}$ is not explicitly used as an output, input/output of the intervener $\phi$ can be presented as $\mathbf{H}_{\texttt{inter}}= \phi(\mathbf{H}_{\texttt{ra}}||\mathbf{H}_{\texttt{env}})$.

\noindent\textbf{FIG-N optimization objective.} The utility loss is computed as $\mathcal{L}_{\texttt{util}}(\mathbf{H}_{\texttt{ra}}||\mathbf{H}_{\texttt{env}}) = \mathcal{L}_{\texttt{task}}(\theta_{\texttt{pred}}\circ\mathtt{Readout}\circ\phi(\mathbf{H}_{\texttt{ra}}||\mathbf{H}_{\texttt{env}}), \mathbf{y})$, where $\mathcal{L}_{\texttt{task}}$ is the task-specific objective. For instance, it could be the mean squared error for regression tasks or the cross-entropy for classification tasks. As introduced in Figure~\ref{fig: rationalization illustration2}, the core of the invariant rationale discovery is to find the graph rationale so that the utility loss attains minimization given \emph{changing environments}. Thus, the total utility objective is
\begin{equation}
\begin{split}
    \mathcal{L}_{\texttt{util}} = \mathcal{L}_{\texttt{util}}(\mathbf{H}_{\texttt{ra}}||\mathbf{H}_{\texttt{env}})+ \alpha\mathcal{L}_{\texttt{util}}(\mathbf{H}_{\texttt{ra}}||\tilde{\mathbf{H}}_{\texttt{env}}),\label{eq: FIG-N utility loss}
\end{split}
\end{equation}
where $\tilde{\mathbf{H}}_{\texttt{env}}$ is the node embeddings from the changing environments. In practical implementations, $\tilde{\mathbf{H}}_{\texttt{env}}$ is the environment node embeddings from other graphs in the mini-batch. Additionally, to fully utilize the rich interactions from the fine-grained intervention module, we apply the following partition regularization term,
\begin{equation}
    \mathcal{L}_{\texttt{reg}}(\mathbf{H}_{\texttt{ra}}||\mathbf{H}_{\texttt{env}}) = \mathbf{s}^{\top}\mathbf{P}(\mathbf{1}-\mathbf{s})
    + (\mathbf{1}-\mathbf{s})^{\top}\mathbf{P}\mathbf{s},\label{eq: regularization term}
\end{equation}
% \hh{intuition: minimizing interaction between rationale and environment, and therefore invariant?} \zhe{Right. Specifically, there are two reasons. (1) Minimizing the rationale-environment interaction. (2) The combination of rationale\&environment is changing (see Eq. (7), which has two rationale-environment combinations), but the groundtruth always comes from the rationale part. In other words, we force the Augmenter to identify which part is truly related with the label (i.e., the rationale). After that, ideally, the model prediction is invariant no matter how the environment part is changing.}
where $\mathbf{P}\in\mathbb{R}^{n\times n}$ is the self-attention matrix from Eq.~\eqref{eq: intervener part 2}, 
% \zhe{rethink the effectiveness of this regularization terms. No interactions between the rationales and environments means they are not similar. If low-similarity leads to better rationale?}
% \yz{Adding furtuer interpretation: ``Namely, the interactions/attentions within the rationale/environment block should be dense, which that between the rationale and environment blocks should be sparse.''} \zhe{It is a good idea to have more elaborations but here if we mentioned some terms like "block", maybe the readers will feel confused about what we are mentioning? If so, let us come up a better elaboration.}\yz{we can simply change block to ``part''}
\begin{equation}
        \mathbf{s}[i]=
        \begin{cases}
            1 & \textrm{if}\  i< K.\\
            0 & \textrm{otherwise}.
        \end{cases}
        \label{eq: FIG-N indicator vector}
\end{equation}
$0$-based indexing is used so there are in total $K$ non-zero entries (i.e., $1$) in $\mathbf{s}$. The meaning of the binary $\mathbf{s}$ vector is to designate whether a particular row of the matrix $\mathbf{H}_{\texttt{ra}}||\mathbf{H}_{\texttt{env}}$ originates from the rationale nodes or the environment nodes. The underlying notion of the regularization term Eq.~\eqref{eq: regularization term} is to impose penalties on interactions between the rationale nodes and the environment nodes. Namely, these two terms $\mathbf{s}^{\top}\mathbf{P}(\mathbf{1}-\mathbf{s})$ and $(\mathbf{1}-\mathbf{s})^{\top}\mathbf{P}\mathbf{s}$ denote the total weights on the links (i.e., cut) between the rationale and environment subgraphs. To handle the changing environments, we introduce an additional regularization term on the changing environments as $\mathcal{L}_{\texttt{reg}}(\mathbf{H}_{\texttt{ra}}||\tilde{\mathbf{H}}_{\texttt{env}})$ where $\tilde{\mathbf{H}}_{\texttt{env}}$ is the environment node embeddings from another graph within the same mini-batch. Then, the total regularization term is
\begin{equation}
    \mathcal{L}_{\texttt{reg}} = \mathcal{L}_{\texttt{reg}}(\mathbf{H}_{\texttt{ra}}||\mathbf{H}_{\texttt{env}})+\mathcal{L}_{\texttt{reg}}(\mathbf{H}_{\texttt{ra}}||\tilde{\mathbf{H}}_{\texttt{env}})\label{eq: node-level regularization},
\end{equation}
and the total objective function is $\mathcal{L}_{\texttt{util}}+\beta\mathcal{L}_{\texttt{reg}}$. To fully harness the capabilities of the fine-grained parametric intervener, it is crucial to note—as highlighted in the introduction—that the behavior of the intervener $\phi$ operates in an adversarial fashion to the other modules. As a result, we formulate a min-max game that involves $\theta=\{\theta_{\texttt{enc}}, \theta_{\texttt{aug-N}},\theta_{\texttt{pred}}\}$ and $\phi$ as,
\begin{equation}
    \min_{\theta}\max_{\phi}\quad \mathcal{L}_{\texttt{util}}+\beta\mathcal{L}_{\texttt{reg}}.\label{eq: final optimization objective}
\end{equation}
Here, the intervener $\phi$ is trained to decrease the utility of the graph rationale by promoting interactions between the rationale nodes and the environment nodes. Conversely, the encoder, augmenter, and predictor (i.e., $\theta$) are optimized in an opposing manner to the intervener's objectives. 

% \zhe{See if it is possible develop a theory for this part to connect to the spectral clustering/partition.}

% \zhe{see if this part is connected to information bottleneck since intuitively $\phi$ will push the $\theta$ to find a minimal sufficient rationale, without involving any changing environments.}

\noindent\textbf{Complexity of FIG-N.} As the encoder $\theta_{\texttt{enc}}$ and the predictor $\theta_{\texttt{pred}}$ are off-the-shelf, the FIG-N introduces two new modules: the node-level augmenter, $\theta_{\texttt{aug-N}}$, and the Transformer-based intervener, $\phi$. Notably, despite these additions, the increase in the number of parameters remains modest. The parameters for $\theta_{\texttt{aug-N}}$ originate from the $\mathtt{MLP}$ defined in Eq.~\eqref{eq: node-level augmenter}. In a configuration where the $\mathtt{MLP}$ has $3$ layers with a feature dimension of $d$, the parameter count is $O(2d^2)$. The intervener $\phi$, driven by the $\mathtt{Transformer}$ layer in Eq.~\eqref{eq: intervener part 1}, has its parameters confined to $O(3d^2+2d^2)=O(5d^2)$, owing to its query, key, value projection matrices and the feed-forward net ($\mathtt{FFN}$ from Eq.~\eqref{eq: feed-forward network}, typically a $3$-layered $\mathtt{MLP}$).

A step-by-step algorithm for FIG-N is in Algorithm~\ref{alg: FIG-N training}. In test phase, the output of $\theta_{\texttt{pred}}\circ\mathtt{Readout}\circ\phi(\mathbf{H}_{\texttt{ra}}||\mathbf{H}_{\texttt{env}})$ is evaluated.

% (Section~\ref{sec: training algorithm}, Appendix)

\begin{algorithm2e}[!t]
\caption{FIG-N single training step for every training graph $\mathcal{G}$}
\label{alg: FIG-N training}
\SetAlgoLined
\SetKwInOut{Input}{Input}\SetKwInOut{Output}{Output}
\Input{a labelled graph $(\mathcal{G}, y)$, a sampled graph $\tilde{\mathcal{G}}$ from the same batch as $\mathcal{G}$, $\theta=\{\theta_{\texttt{enc}}, \theta_{\texttt{aug-N}},\theta_{\texttt{pred}}\}$, $\phi$;}
\Output{updated $\theta$ and $\phi$;}

compute $\mathbf{H}=\theta_{\texttt{enc}}(G)$ and $\tilde{\mathbf{H}}=\theta_{\texttt{enc}}(\tilde{G})$\;
compute $(\mathbf{H}_{\texttt{ra}}, \mathbf{H}_{\texttt{env}})=\theta_{\texttt{aug-N}}(\mathbf{H})$, $(\tilde{\mathbf{H}}_{\texttt{ra}}, \tilde{\mathbf{H}}_{\texttt{env}})=\theta_{\texttt{aug-N}}(\tilde{\mathbf{H}})$ via Eqs.~\eqref{eq: node-level augmenter},~\eqref{eq: rationale node embeddings}, and~\eqref{eq: environment node embeddings}\;
concatenate rationale-environment pairs $\mathbf{H}_{\texttt{ra}}||\mathbf{H}_{\texttt{env}}$ and $\mathbf{H}_{\texttt{ra}}||\tilde{\mathbf{H}}_{\texttt{env}}$\;
compute $\mathcal{L}_{\texttt{util}}$ via Eq.~\eqref{eq: FIG-N utility loss}\;
compute $\mathcal{L}_{\texttt{reg}}$ via Eqs.~\eqref{eq: node-level regularization} and~\eqref{eq: FIG-N indicator vector}\;
% compute gradients $\frac{\partial (\mathcal{L}_{\texttt{util}}+\beta\mathcal{L}_{\texttt{reg}})}{\partial \theta}$ and $\frac{\partial (\mathcal{L}_{\texttt{util}}+\beta\mathcal{L}_{\texttt{reg}})}{\partial \phi}$\;
update $\theta$ via gradient descent with $\frac{\partial (\mathcal{L}_{\texttt{util}}+\beta\mathcal{L}_{\texttt{reg}})}{\partial \theta}$\;
update $\phi$ via gradient ascent with $\frac{\partial (\mathcal{L}_{\texttt{util}}+\beta\mathcal{L}_{\texttt{reg}})}{\partial \phi}$\;
\end{algorithm2e}

% The only extra module we introduced in FIG-N compared to an off-the-shelf graph Transformer $f_{\theta}$ is the node partitioner $g_{\phi}$. As we mentioned in the introduction of $g_{\phi}$ (Eq.~\eqref{eq: node partitioner}, we select $g_{\phi}$ to be a light-weight GraphTrans~\cite{DBLP:journals/corr/abs-2201-08821}. In our experiments we set the $g_{\phi}$ to be only $1$-layered GIN + $1$-layered single-head Transformer for efficiency, whose total number of parameters is from $7$ linear feature transformations, i.e., $O(7d^2)$ (2 from GIN~\cite{DBLP:conf/iclr/XuHLJ19}, 3 from Transformer, and 2 from FFN).

\subsection{Virtual Node-Level Variant: FIG-VN}
In the previously introduced FIG-N, its augmenter decomposes the nodes into rationale nodes and environment nodes via a trainable node partitioner $\theta_{\texttt{aug-N}}$ so that the interaction is conducted at the node level (Eqs.~\eqref{eq: intervener part 1}) whose dense attention matrix's complexity is quadratic in terms of the node numbers, infeasible for large graphs. This section extends this idea to extract the graph rationale at the virtual node level, which has a lower computation complexity compared to FIG-N and provides an intermediate intervention granularity between the node-level model (FIG-N) and the graph-level model (GREA~\cite{DBLP:conf/kdd/LiuZXL022}).

\noindent\textbf{Virtual node-level augmenter.} Our idea is partly inspired by the speedup technique from Linformer~\cite{DBLP:journals/corr/abs-2006-04768}, which reformulates both the attention matrix and node (token) embedding matrix to dimensions of $\mathbb{R}^{n\times r}$ and $\mathbb{R}^{r\times d}$, respectively. This reformulation ensures that their multiplication scales linearly with the number of nodes (tokens) $n$. Within this configuration, $r$, a pre-defined rank, is significantly smaller than $n$, and $d$ represents the feature dimension. Drawing from the Linformer technique, we propose that the restructured token embedding matrix, with dimensions of $\mathbb{R}^{r\times d}$, can be interpreted as embeddings for $r$ virtual nodes.

Building upon this insight, given node embeddings $\mathbf{H}$ from the encoder, the virtual node embeddings are:
\begin{equation}
    \mathbf{H}_{\texttt{VN}} = \mathtt{Softmax}(\mathbf{W}_{\texttt{N-VN}})\mathbf{H}.
    \label{eq: virtual node embeddings}
\end{equation}
Here, the row-wise applied $\mathtt{Softmax}$ function, along with\\ $\mathtt{Softmax}(\mathbf{W}_{\texttt{N-VN}}) \in \mathbb{R}^{r\times n}$, yields a trainable matrix assigning $n$ nodes to $r$ virtual nodes, where $r$ acts as a tunable hyper-parameter. In experiments, we set $r=8$. As all the virtual node embeddings are learned, a subset of the $r$ virtual nodes can be designated as rationale virtual nodes, whose rationality is data-driven by the intervention procedure discussed in subsequent subsections. For brevity, the initial $K$ virtual nodes are deemed as rationale virtual nodes, while the last $r-K$ nodes is considered the environment virtual node. Like the FIG-N, here $K$ is a hyperparameter whose impact is studied in Section~\ref{sec: sensitivity study}. Thus, rationale and environment embeddings are presented as:
\begin{subequations}
\begin{align}
\mathbf{H}_{\texttt{ra}}&=\mathbf{H}_{\texttt{VN}}[:K, :]\in\mathbb{R}^{K\times d},\label{eq: rationale virtual node embeddings FIG-VN}\\
\mathbf{H}_{\texttt{env}}&=\mathbf{H}_{\texttt{VN}}[K:, :]\in\mathbb{R}^{(r-K)\times d}.\label{eq: environment virtual node embeddings FIG-VN}
\end{align}
\end{subequations}
% For the virtual node-level augmenter $\theta_{\texttt{aug-VN}}$, the rationale/environment decomposition remains non-parametric. 
The parameter of $\theta_{\texttt{aug-VN}}$ is $\mathbf{W}_{\texttt{N-VN}}$.

\noindent\textbf{Virtual node-level intervener.} This section discusses the design of a virtual node-level intervener, which parallels the framework presented in Section~\ref{sec: FIG-N}. The salient difference lies in that the intervention here functions on the virtual nodes rather than the given real nodes. Building upon our previous steps, we obtain the rationale virtual node embeddings, $\mathbf{H}_{\texttt{ra}}\in\mathbb{R}^{K\times d}$, and the environment node embedding, $\mathbf{H}_{\texttt{env}}\in\mathbb{R}^{(r-K)\times d}$. Thanks to the property of the Transformer that it can process sets with variable size, the design of the virtual node-level intervener $\phi$ is similar to the node-level intervener as $\mathbf{H}_{\texttt{inter}}, \mathbf{P} = \mathtt{Transformer}(\mathbf{H}_{\texttt{ra}}||\mathbf{H}_{\texttt{env}})$ or short as $\mathbf{H}_{\texttt{inter}}=\phi(\mathbf{H}_{\texttt{ra}}||\mathbf{H}_{\texttt{env}})$ if the attention matrix $\mathbf{P}$ is not used. Notably, for FIG-VN,  $\mathbf{P}\in\mathbb{R}^{r\times r}$ describes the interaction among the $r$ virtual nodes.

% \begin{equation}
%     \phi_{\texttt{inter-VN}}(\mathbf{H}_{\texttt{ra}}||\mathbf{H}_{\texttt{env}})= \mathbf{P}\mathbf{H}_{\texttt{inter}}+\mathbf{H}_{\texttt{inter}},
% \end{equation}
% with $\mathbf{H}_{\texttt{inter}}=\mathbf{H}_{\texttt{ra}}||\mathbf{H}_{\texttt{env}}$ and $\mathbf{P}= \mathtt{Attn}(\mathbf{\mathbf{H}_{\texttt{inter}}})$. 

\noindent\textbf{FIG-VN optimization objective.} The output of $\theta_{\texttt{pred}}\circ\mathtt{Readout}\circ\phi(\cdot)$ is used for minimizing the utility loss $\mathcal{L}_{\texttt{util}} = \mathcal{L}_{\texttt{util}}(\mathbf{H}_{\texttt{ra}}||\mathbf{H}_{\texttt{env}})+\alpha\mathcal{L}_{\texttt{util}}(\mathbf{H}_{\texttt{ra}}||\tilde{\mathbf{H}}_{\texttt{env}})$, where $\mathcal{L}_{\texttt{util}}(\mathbf{H}_{\texttt{ra}}||\mathbf{H}_{\texttt{env}}) = \mathcal{L}_{\texttt{task}}(\theta_{\texttt{pred}}\circ\mathtt{Readout}\circ\phi(\mathbf{H}_{\texttt{ra}}||\mathbf{H}_{\texttt{env}}), \mathbf{y})$ and $\mathcal{L}_{\texttt{util}}(\mathbf{H}_{\texttt{ra}}||\tilde{\mathbf{H}}_{\texttt{env}})$ is defined similarly. For modeling the changing environment, $\tilde{\mathbf{H}}_{\texttt{env}}$ is the virtual node embeddings from other graphs in the mini-batch. Additionally, the previously proposed regularization term Eq,~\eqref{eq: regularization term} can be extended to the virtual node-level variant: $\mathcal{L}_{\texttt{reg}}(\mathbf{H}_{\texttt{ra}}||\mathbf{H}_{\texttt{env}}) = \mathbf{s}^{\top}\mathbf{P}(\mathbf{1}-\mathbf{s})+ (\mathbf{1}-\mathbf{s})^{\top}\mathbf{P}\mathbf{s}$. The total regularization term, considering the changing environment $\tilde{\mathbf{H}}_{\texttt{env}}$, is $\mathcal{L}_{\texttt{reg}} = \mathcal{L}_{\texttt{reg}}(\mathbf{H}_{\texttt{ra}}||\mathbf{H}_{\texttt{env}})+\mathcal{L}_{\texttt{reg}}(\mathbf{H}_{\texttt{ra}}||\tilde{\mathbf{H}}_{\texttt{env}})$. As the $\mathbf{P}$ depicts interactions among virtual nodes, we construct the rationale/environment indicator vector $\mathbf{s}$ analogously to Eq.~\eqref{eq: FIG-N indicator vector}.
% , as:
% \begin{equation}
%         \mathbf{s}[i]=
%         \begin{cases}
%             1 & if\  i< r,\\
%             0 & else.
%         \end{cases}
%         \label{eq: FIG-VN indicator vector}
% \end{equation}
% Here, the initial $r-1$ entries of $\mathbf{m}{\texttt{inter}}$ are set to $1$, marking the first $r-1$ virtual nodes as graph rationales, with the last entry ($0$) indicating the final virtual node as the environment. 
Put everything together, and the optimization objective of FIG-VN is $\min_{\theta}\max_{\phi}\  \mathcal{L}_{\texttt{util}}+\beta\mathcal{L}_{\texttt{reg}}$, where $\theta=\{\theta_{\texttt{enc}}, \theta_{\texttt{aug-VN}},\theta_{\texttt{pred}}\}$.

\noindent\textbf{Complexity of FIG-VN.} As we mentioned the encoder $\theta_{\texttt{enc}}$ and the predictor $\theta_{\texttt{pred}}$ are off-the-shelf. Thus, the extra modules introduced by the FIG-VN are the virtual node-level augmenter $\theta_{\texttt{aug-VN}}$ and the Transformer-based intervener $\phi$. The parameters for $\theta_{\texttt{aug-VN}}$ originate from the matrix $\mathbf{W}_{\texttt{N-VN}}$, as defined in Eq.~\eqref{eq: virtual node embeddings}. The number of these parameters is in order $O(nr)$, where $n$ denotes the number of nodes. For practical implementation purposes, $n$ is pre-set; it is set to $10\times$ average size of graphs from the dataset, and we truncate the input graphs if its size is larger than $10\times$ average size. The intervener $\phi$ parameters originate from the $\mathtt{Transformer}$ layer, outlined in Eq.~\eqref{eq: intervener part 1}. The number of parameters here is $O(5d^2)$, owing to its query, key, value projection matrices, and the feed-forward net (Eq.~\eqref{eq: feed-forward network}, typically a $3$-layered $\mathtt{MLP}$).

% Compared to FIG-N, both interveners share the same number of parameters, but the augmenter of virtual node-level intervener $\theta_{\texttt{aug-VN}}$ has less parameters ($O(nr)$) than the node-level intervener $\theta_{\texttt{aug-N}}$ ($O(d^2)$). 

A step-by-step algorithm for FIG-VN is in Algorithm~\ref{alg: FIG-VN training}, Appendix. In test phase, the output of $\theta_{\texttt{pred}}\circ\mathtt{Readout}\circ\phi(\mathbf{H}_{\texttt{ra}}||\mathbf{H}_{\texttt{env}})$ is evaluated.

\section{Experiments}
\label{sec: experiments}

\begin{table*}[th!]
\centering
\caption{Effectiveness comparison (mean$\pm$std) with baseline methods. ($\downarrow$) denotes the lower the better and  ($\uparrow$) denotes the higher the better. \colorbox{lightlightgray}{Statistics in grey}are reported in the original papers. The best is bold, and the second best is underlined. N/A means the method cannot work on regression tasks.}
\label{tab: effectiveness comparison}
\resizebox{.75\linewidth}{!}{
\begin{tabular}{l|lll|llll}
\toprule
 & \multicolumn{3}{c|}{Graph Regression} & \multicolumn{4}{c}{Graph Classification} \\
Dataset & ZINC & AQSOL & mollipo & molhiv & moltox21 & molbace & molbbbp \\
Metric & MAE($\downarrow$) & MAE($\downarrow$) & RMSE($\downarrow$) & AUC($\uparrow$) & AUC($\uparrow$) & AUC($\uparrow$) & AUC($\uparrow$) \\\midrule
GIN & 0.350\scriptsize$\pm$0.008 & 1.237\scriptsize$\pm$0.011 & 0.783\scriptsize$\pm$0.017 & \cellcolor[HTML]{D9D9D9}77.1\scriptsize$\pm$1.5 & 75.6\scriptsize$\pm$0.9 & 80.7\scriptsize$\pm$1.2 & 69.5\scriptsize$\pm$1.0 \\
GAT & 0.723\scriptsize$\pm$0.010 & 1.638\scriptsize$\pm$0.048 & 0.923\scriptsize$\pm$0.011 & 75.0\scriptsize$\pm$0.5 & 72.2\scriptsize$\pm$0.6 & 75.3\scriptsize$\pm$0.8 & 67.1\scriptsize$\pm$0.6 \\
GATv2 & 0.729\scriptsize$\pm$0.015 & 1.722\scriptsize$\pm$0.022 & 0.943\scriptsize$\pm$0.021 & 72.2\scriptsize$\pm$0.5 & 73.6\scriptsize$\pm$0.2 & 76.8\scriptsize$\pm$1.6 & 65.7\scriptsize$\pm$0.7 \\
GatedGCN & 0.579\scriptsize$\pm$0.023 & 1.533\scriptsize$\pm$0.035 & 0.819\scriptsize$\pm$0.033 & 74.8\scriptsize$\pm$1.6 & 75.0\scriptsize$\pm$0.8 & 81.2\scriptsize$\pm$1.2 & 68.3\scriptsize$\pm$0.9 \\\midrule
GT & \cellcolor[HTML]{D9D9D9}0.226\scriptsize$\pm$0.014 & 1.319\scriptsize$\pm$0.026 & 0.882\scriptsize$\pm$0.020 & 73.5\scriptsize$\pm$0.4 & 75.0\scriptsize$\pm$0.6 & 77.1\scriptsize$\pm$2.3 & 65.0\scriptsize$\pm$1.1 \\
GraphiT & \cellcolor[HTML]{D9D9D9}0.202\scriptsize$\pm$0.011 & 1.162\scriptsize$\pm$0.005 & 0.846\scriptsize$\pm$0.023 & 74.6\scriptsize$\pm$1.0 & 71.8\scriptsize$\pm$1.3 & 73.4\scriptsize$\pm$3.6 & 64.6\scriptsize$\pm$0.5 \\
SAN & \cellcolor[HTML]{D9D9D9}0.139\scriptsize$\pm$0.006 & 1.199\scriptsize$\pm$0.218 & 0.816\scriptsize$\pm$0.112 & \cellcolor[HTML]{D9D9D9}77.9\scriptsize$\pm$0.2 & 71.3\scriptsize$\pm$0.8 & 79.0\scriptsize$\pm$3.1 & 63.8\scriptsize$\pm$0.9 \\
SAT & \cellcolor[HTML]{D9D9D9}0.094\scriptsize$\pm$0.008 & 1.236\scriptsize$\pm$0.023 & 0.835\scriptsize$\pm$0.008 & 78.8\scriptsize$\pm$0.6 & 75.6\scriptsize$\pm$0.7 & 83.6\scriptsize$\pm$2.1 & 69.6\scriptsize$\pm$1.3 \\
Graphormer & \cellcolor[HTML]{D9D9D9}0.122\scriptsize$\pm$0.006 & 1.265\scriptsize$\pm$0.025 & 0.911\scriptsize$\pm$0.015 & 79.3\scriptsize$\pm$0.4 & 77.3\scriptsize$\pm$0.8 & 79.3\scriptsize$\pm$3.0 & 67.7\scriptsize$\pm$0.9 \\
GraphTrans & 0.192\scriptsize$\pm$0.011 & 1.233\scriptsize$\pm$0.052 & 0.915\scriptsize$\pm$0.032 & 78.1\scriptsize$\pm$0.5 & 76.4\scriptsize$\pm$0.8 & 78.0\scriptsize$\pm$1.8 & 70.5\scriptsize$\pm$0.9 \\
GPS & \cellcolor[HTML]{D9D9D9}\textbf{0.070\scriptsize$\pm$0.004} & 1.032\scriptsize$\pm$0.007 & 0.780\scriptsize$\pm$0.021 & \cellcolor[HTML]{D9D9D9}78.8\scriptsize$\pm$1.0 & 75.7\scriptsize$\pm$0.4 & 79.6\scriptsize$\pm$1.4 & 69.6\scriptsize$\pm$1.1 \\\midrule
DIR & N/A & N/A & N/A & \cellcolor[HTML]{D9D9D9}77.1\scriptsize$\pm$0.6 & 73.1\scriptsize$\pm$0.2 & 74.8\scriptsize$\pm$0.3 & 70.5\scriptsize$\pm$1.4 \\
GREA & 0.227\scriptsize$\pm$0.020 & 1.177\scriptsize$\pm$0.019 & 0.769\scriptsize$\pm$0.025 & \cellcolor[HTML]{D9D9D9}79.3\scriptsize$\pm$0.9 & \cellcolor[HTML]{D9D9D9}{\ul 78.2\scriptsize$\pm$0.9} & \cellcolor[HTML]{D9D9D9}82.4\scriptsize$\pm$2.4 & \cellcolor[HTML]{D9D9D9}69.9\scriptsize$\pm$1.8 \\\midrule
FIG-N & 0.095\scriptsize$\pm$0.008 & \textbf{ 0.990\scriptsize$\pm$0.012} & {\ul0.708\scriptsize$\pm$0.013} & {\ul 80.1\scriptsize$\pm$0.7} & \textbf{78.8\scriptsize$\pm$0.5} & \textbf{85.3\scriptsize$\pm$2.0} & \textbf{73.8\scriptsize$\pm$0.7} \\
FIG-VN & {\ul 0.086\scriptsize$\pm$0.012} & {\ul1.011\scriptsize$\pm$0.009} & \textbf{0.706\scriptsize$\pm$0.009} & \textbf{80.2\scriptsize$\pm$1.0} & {\ul78.2\scriptsize$\pm$0.6} & {\ul84.5\scriptsize$\pm$1.3} & {\ul 73.1\scriptsize$\pm$0.8}\\
\bottomrule
\end{tabular}
}
\end{table*}

In this section, we begin by detailing the datasets and baseline methods. Subsequent subsections evaluate the effectiveness of our approach, supplemented by an efficiency study, an ablation study, a sensitivity study, a visualization of the attention matrix, and a visualization of the rationales detected. Convergence analysis can be found in Appendix, Section~\ref{sec: training convergence}.

\subsection{Setup}

In this paper, we use $7$ publicly-available real-world datasets: (1) graph classification datasets molhiv~\cite{DBLP:journals/corr/abs-2005-00687}, moltox21~\cite{DBLP:journals/corr/abs-2005-00687}, molbace~\cite{DBLP:journals/corr/abs-2005-00687}, molbbbp~\cite{DBLP:journals/corr/abs-2005-00687} and (2) graph regression datasets ZINC~\cite{DBLP:journals/jmlr/DwivediJL0BB23}, AQSOL~\cite{DBLP:journals/jmlr/DwivediJL0BB23}, and mollipo~\cite{DBLP:journals/corr/abs-2005-00687}. We strictly follow the metrics and dataset split recommended by the given benchmarks. To be concrete, area under the ROC curve (AUC) is the metric for datasets molhiv, moltox21, molbace, molbbbp; root-mean-square deviation (RMSE) is the metric for dataset mollipo; mean absolute error (MAE) is the metric of datasets ZINC and AQSOL. The detailed statistics of the datasets are given in Table~\ref{tab: dataset statistics} (Appendix). We report the average result with the standard deviation in $10$ runs.

Our baseline methods include (1) $4$ graph neural network models: GIN~\cite{DBLP:conf/iclr/XuHLJ19}, GAT~\cite{DBLP:conf/iclr/VelickovicCCRLB18}, GATv2~\cite{DBLP:conf/iclr/Brody0Y22}, and GatedGCN~\cite{DBLP:journals/corr/abs-1711-07553} (2) $7$ graph Transformers: GT~\cite{DBLP:journals/corr/abs-2012-09699}, GraphiT~\cite{DBLP:journals/corr/abs-2106-05667}, SAN~\cite{DBLP:conf/nips/KreuzerBHLT21}, SAT~\cite{DBLP:conf/icml/ChenOB22}, Graphormer~\cite{DBLP:conf/nips/YingCLZKHSL21}, GraphTrans~\cite{DBLP:journals/corr/abs-2201-08821}, GPS~\cite{DBLP:conf/nips/RampasekGDLWB22}, and (3) $2$ graph rationale discovery methods: DIR~\cite{DBLP:conf/iclr/WuWZ0C22} and GREA~\cite{DBLP:conf/kdd/LiuZXL022}.

% possible baselines: Cooperative Classification and Rationalization for Graph Generalization, Unleashing the Power of Graph Data Augmentation on Covariate Distribution Shift, Invariant Graph Learning for Causal Effect Estimation, Empowering Graph Invariance Learning with Deep Spurious Infomax, How Interpretable Are Interpretable Graph Neural Networks?

\begin{table}[t!]
\centering
\caption{Number of parameter and FLOPs comparisons between the proposed augmenter, intervener, and common graph encoders.}
\label{tab: number of parameters}
\resizebox{.7\linewidth}{!}{
\begin{tabular}{lll}
\toprule
Model & \# Parameters & FLOPs\\\midrule
GIN & $1,708,807$ & $53,008,220$\\
SAT & $2,790,739$ & $101,520,116$\\
GraphTrans & $2,793,307$ & $111,548,906$\\
GPS & $3,236,239$ & $133,229,235$\\
$\{\theta_\texttt{aug-N},\phi\}$ & $453,001$ & $31,303,800$\\
$\{\theta_\texttt{aug-VN},\phi\}$ & $363,320$ & $14,558,400$\\
\bottomrule
\end{tabular}
}
\end{table}

\subsection{Effectiveness Study}
The effectiveness comparison between the proposed FIG-N, FIG-VN, and baseline methods are provided in Table~\ref{tab: effectiveness comparison}. To ensure a fair comparison, certain pre-trained models, such as the pre-trained Graphormer~\cite{DBLP:conf/nips/YingCLZKHSL21}, are omitted. As DIR~\cite{DBLP:conf/iclr/WuWZ0C22} is designed to conduct interventions on the label prediction vectors, it cannot be directly applied to graph regression tasks.

We have several observations. First, our proposed FIG-N and FIG-VN consistently outperform, or are at least on par with, all the baseline methods on the graph classification and regression datasets. Second, the virtual-node variant FIG-VN does not lead to significant performance degradation compared to the node-level variant FIG-N.

\subsection{Efficiency Study}
\label{sec: efficiency study}

A detailed comparison regarding the number of parameters and the FLOPs (floating point operations) is presented in Table~\ref{tab: number of parameters}, where we list $4$ typical $5$-layered encoders (GIN, SAT, GraphTrans, and GPS), and our proposed node-/virtual node-level augmenter, intervener modules (i.e., $\{\theta_\texttt{aug-N},\phi\}$ and $\{\theta_\texttt{aug-VN},\phi\}$). The comparison shows that our proposed interveners are lightweight and only increase very minor computational costs.

We also compare the model efficiency (training iterations/second) of FIG-N and FIG-VN in Table~\ref{tab: wall-clock time comparison} working with different encoders (GIN and GPS). The batch size is set as $32$. We note that the inclusion of our proposed augmenter and intervener, represented as $\{\theta_\texttt{aug-N},\phi\}$ or $\{\theta_\texttt{aug-VN},\phi\}$, introduces a slight reduction in training speed. That is because the proposed parametric augmenter and intervener increase the steps of the data pipeline, as presented in figure~\ref{fig: FIG pipeline}, and enlarge the computational graph for auto-gradient tools, such as PyTorch. Fortunately, the parameter count of the parametric augmenter and intervener is low, ensuring that the overall training speed of the model is not dramatically affected.

% A detailed comparison about the number of parameters of the newly-introduced node-/virtual node-level augmenter ($\theta_{\texttt{aug-N}}$/$\theta_{\texttt{aug-VN}}$), intervener ($\phi$) modules and other typical encoder modules ($\theta_{\texttt{enc}}$) are presented in Table~\ref{tab: number of parameters}.

\begin{table}[t!]
\centering
\caption{Wall-clock time (iterations/second) comparison of different encoder-intervener combinations. The larger, the faster. ($\downarrow$) denotes the speed degradation compared with the vanilla encoder.}
\label{tab: wall-clock time comparison}
\resizebox{.9\linewidth}{!}{
\begin{tabular}{ll|lll}
\toprule
Encoder & Intervener & mollipo & molbace & molbbbp \\\midrule
\multirow{3}{*}{GIN} & None & 29.38 & 25.62 & 27.11 \\
 & FIG-N & 23.05\scriptsize($\downarrow$6.33) & 21.23\scriptsize($\downarrow$4.39) & 21.61\scriptsize($\downarrow$5.50) \\
 & FIG-VN & 23.35\scriptsize($\downarrow$6.03) & 21.46\scriptsize($\downarrow$4.16) & 22.32\scriptsize($\downarrow$4.79) \\\midrule
\multirow{3}{*}{GPS} & None & 24.29 & 20.51 & 22.30 \\
 & FIG-N & 19.57\scriptsize($\downarrow$4.72) & 17.83\scriptsize($\downarrow$2.68) & 18.67\scriptsize($\downarrow$3.63) \\
 & FIG-VN & 19.93\scriptsize($\downarrow$4.63) & 18.16\scriptsize($\downarrow$2.35) & 18.88\scriptsize($\downarrow$3.42)\\\bottomrule
\end{tabular}
}
\end{table}

\subsection{Ablation Study}
We conducted an ablation study on the proposed models, FIG-N and FIG-VN. We designed two ablated variants as baselines: (1) $\theta_{\texttt{enc}}\circ\theta_{\texttt{pred}}$ which is a pure composition of the encoder $\theta_{\texttt{enc}}$ and the predictor $\theta_{\texttt{pred}}$ without any rationale discovery module. Many of the existing graph classifiers are in this form, and here we select the GraphGPS~\cite{DBLP:conf/nips/RampasekGDLWB22}, which is also the backbone of our FIG model. (2) FIG-N w/o reg and FIG-VN w/o reg which remove the regularization term (Eq.~\eqref{eq: node-level regularization}) from the objective function. Our results in Table~\ref{tab: ablation study} highlight that (1) equipped with the proposed Transformer-based intervener, the model's performance improves across all the datasets; e.g., the AUC is improved from $79.6\%$ to $84.3\%$ (FIG-N) on the molbace dataset. (2) With the proposed regularization term, the model's performance can be improved further; e.g., the AUC of the FIG-N is improved from $72.3\%$ to $73.8\%$ on the molbbbp dataset.

\begin{table}[t!]
\centering
\caption{Ablation study (mean$\pm$std) of the proposed model FIG. ($\downarrow$) denotes the lower the better and  ($\uparrow$) denotes the higher the better.}
\label{tab: ablation study}
\resizebox{.8\columnwidth}{!}{
\begin{tabular}{l|lll}
\toprule
Dataset & mollipo & molbace & molbbbp \\
Metric & RMSE($\downarrow$) & AUC($\uparrow$) & AUC($\uparrow$) \\\midrule
$\theta_{\texttt{enc}}\circ\theta_{\texttt{pred}}$ & 0.780\scriptsize$\pm$0.021 & 79.6\scriptsize$\pm$1.4 & 70.5\scriptsize$\pm$0.9 \\\midrule
FIG-N w/o reg & 0.736\scriptsize$\pm$0.022 & 84.3\scriptsize$\pm$0.7 & 72.3\scriptsize$\pm$1.0 \\
FIG-VN w/o reg & 0.758\scriptsize$\pm$0.018 & 83.2\scriptsize$\pm$1.5 & 71.9\scriptsize$\pm$0.7 \\\midrule
FIG-N w/ reg & 0.708\scriptsize$\pm$0.013 & 85.3\scriptsize$\pm$2.0 & 73.8\scriptsize$\pm$0.7 \\
FIG-VN w/ reg & 0.706\scriptsize$\pm$0.009 & 84.5\scriptsize$\pm$1.3 & 73.1\scriptsize$\pm$0.8\\
\bottomrule
\end{tabular}
}
\end{table}

\section{Related Work}
\label{sec: related work}

This section introduces two topics related to this paper: (1) invariant learning on graphs and (2) Transformer on graphs.
\noindent\textbf{Invariant Learning on Graphs.} Invariant learning, which has extensive applications in many fields such as computer vision~\cite{DBLP:journals/corr/abs-2206-15475,DBLP:journals/corr/abs-1907-02893,DBLP:conf/icml/GaninL15,DBLP:conf/icml/ChangZYJ20}, is gaining more attention in the community of graph machine learning. OOD-GNN~\cite{DBLP:journals/tkde/LiWZZ23}, for instance, applies random Fourier features to eliminate the statistical dependence between relevant and irrelevant graph representations.~\cite{DBLP:conf/icml/BevilacquaZ021} studies the graph invariant representation learning with a particular focus on the size discrepancies between the training/test graphs. DIR~\cite{DBLP:conf/iclr/WuWZ0C22} decomposes the given graph into a rationale subgraph and an environment subgraph; after that, it uses a graph classifier to conduct prediction on the above two graphs respectively and its intervention is via the Hadamard product between the prediction vectors. Similarly, GREA~\cite{DBLP:conf/kdd/LiuZXL022} conducts the rationale/environment decomposition at the node level and its intervention operation is to directly add the environment graph embedding into the rationale graph embedding. In a similar vein, GIL~\cite{DBLP:conf/nips/LiZ0022} decomposes the given graph into the invariant and variant subgraphs; based on that, it infers the environment label, as input of the invariance regularization term~\cite{DBLP:journals/corr/abs-2008-01883}. Furthermore, invariant learning can also benefit graph contrastive learning~\cite{DBLP:conf/www/SuiTCFGCLZW24} to enhance data augmentation. Using invariant learning to address the OOD generalization challenges on graph~\cite{DBLP:journals/corr/abs-2202-07987,DBLP:conf/nips/GuiLWJ22,DBLP:conf/www/YueLLGYL24,DBLP:conf/nips/SuiWWCLZW023,DBLP:conf/icml/Yao0C0S024} is promising, but our paper does not concentrate on this setting, and we leave it as the future work.

% DBLP:conf/icml/LiWZW0C22,

\noindent\textbf{Transformer on Graphs.} The Transformer architecture~\cite{DBLP:conf/nips/VaswaniSPUJGKP17} has achieved significant success in various domains, including natural language processing~\cite{DBLP:conf/nips/VaswaniSPUJGKP17}, computer vision~\cite{DBLP:journals/pami/00020C0GLTXXXYZ23}, and more~\cite{DBLP:journals/corr/abs-2206-06488}. In recent years, there has been a surge of interest in enhancing graph machine learning methods by leveraging the capabilities of Transformers. For example, GraphTrans~\cite{DBLP:journals/corr/abs-2201-08821} concatenates the Transformer architecture after various message-passing neural networks; GPS~\cite{DBLP:conf/nips/RampasekGDLWB22} proposes a powerful layer which operates both a graph convolution layer and a Transformer layer in parallel; GT~\cite{DBLP:journals/corr/abs-2012-09699} generalizes the attention module on graphs via concatenating the spectral vectors with the raw node features and computing the attention weights on all the existing edges; Graphormer~\cite{DBLP:conf/nips/YingCLZKHSL21} encodes both the node centrality and node-pair shortest path into the Transformer architecture. From the architecture perspective, a comprehensive survey about graph Transformer can be found in~\cite{DBLP:journals/corr/abs-2202-08455}.

\section{Conclusion}
\label{sec: conclusion}
This paper studies the invariant rationale discovery problem on graph data. Distinct from existing methods, our proposed solutions (FIG-N and FIG-VN) are designed at more fine-grained levels, specifically at the node and virtual node levels, so that they can better model the interactions between the rationale and environment subgraphs. More importantly, we formulate the intervener and the other model modules in a min-max game, which can significantly improve the quality of the extracted graph rationales. Comprehensive experiments on $7$ real-world datasets illustrate the effectiveness of the proposed method compared to $13$ baseline methods. Our evaluation of the proposed augmenter's and intervener's efficiency shows that they can largely retain the overall training efficiency.

% \section{Limitation and Future Work}
% The proposed FIG's main limitation is the increased number of parameters, which is inevitable due to the intervener's parameterizing. The computational graph expands since the intervener is appended to the existing encoder and predictor modules. For auto-gradient tools, such as PyTorch, this results in an increase in the training duration. Our future work includes employing the proposed FIG for OOD generalization and model explainability.

% \section{Broader Impact}
% This paper presents work whose goal is to advance the field of Machine Learning. There are many potential societal consequences of our work, none of which we feel must be specifically highlighted here.

\clearpage

\bibliographystyle{named}
\bibliography{ref}

\clearpage
\appendix

\onecolumn
\begin{center}
    \LARGE Appendix of `Fine-grained Graph Rationalization'
\end{center}

\section{Dataset Statistics}
The statistics of the datasets are presented in Table~\ref{tab: dataset statistics}.

\begin{table*}[th]
\centering
\caption{Dataset statistics.}
\label{tab: dataset statistics}
\begin{tabular}{cccccccc}
\toprule
\textbf{Name} & \textbf{\# Graphs} & \textbf{\# Nodes} & \textbf{\# Edges} & \textbf{\# Features} & \textbf{\# Classes} & \textbf{Split}     & \textbf{Metric} \\\midrule
ZINC          & 12000              & 23.2              & 49.8              & 21 (node), 4 (edge)   & N/A                 & 10000/1000/1000    & MAE             \\
AQSOL         & 9833               & 17.6              & 35.8              & 65 (node), 5 (edge)  & N/A                 & 7836/998/999       & MAE             \\
mollipo       & 4200               & 27.0              & 59.0              & 9 (node), 3 (edge)   & N/A                 & 3360/420/420       & RMSE             \\
molhiv        & 41127              & 25.5              & 54.9              & 9 (node), 3 (edge)   & 2                   & 32901/4113/4113    & AUC         \\
% molpcba       & 497929             & 26.0              & 56.2              & 9 (node), 3 (edge)   & 2                   & 350343/43793/43793 & AP              \\
moltox21      & 7831               & 18.6              & 38.6              & 9 (node), 3 (edge)   & 2                   & 6264/783/784       & AUC         \\
molbace       & 1513       & 34.4   & 73.7  &  9 (node), 3 (edge) & 2 & 1210/151/152 & AUC\\
molbbbp       & 2039       & 24.1   & 51.9  &  9 (node), 3 (edge) & 2 & 1631/204/204 & AUC\\
molmuv       & 93087       & 24.2   & 52.6  &  9 (node), 3 (edge) & 2 & 74469/9309/9309 & AUC\\ 
\bottomrule
\end{tabular}
\end{table*}

\section{Hardware and Implementations.}
\label{sec: hardware and implmentations}
We implement FIG-N, FIG-VN, and all the baseline methods in PyTorch\footnote{\url{https://pytorch.org/}} and PyTorch-geometric\footnote{\url{https://pytorch-geometric.readthedocs.io/en/latest/}}. All the efficiency study results are from one NVIDIA Tesla V100 SXM2-32GB GPU on a server with 96 Intel(R) Xeon(R) Gold 6240R CPU @ 2.40GHz processors and 1.5T RAM. We directly use those statistics when baseline methods have pre-existing results for specific datasets. In cases where such results are absent, we implement the models based on either the available code or details described in the associated publications.

We first introduce some shared implementations among FIG-N/VN and baseline methods. The batch size is set as $32$, and the weight decay is set as $0$. The hidden dimension is set as $300$. The dropout rate for both the encoder and the intervener is searched between $\{0, 0.5\}$. The pooling is searched between $\{\texttt{mean}, \texttt{sum}\}$. The normalization is searched between the batch normalization~\cite{DBLP:conf/icml/IoffeS15} and layer normalization~\cite{DBLP:journals/corr/BaKH16}. The learning rate is initialized as $0.0001$ and will decay by a factor $\frac{1}{4}$ if the validation performance is not improved in $10$ epochs. Next, we detail the implementation of FIG-N/VN and baseline methods.

\subsection{Implementation of FIG-N/VN}
The encoder is set as GPS~\cite{DBLP:conf/nips/RampasekGDLWB22} on ZINC, AQSOL, mollipo, molhiv, molbace, and set as GraphTrans~\cite{DBLP:journals/corr/abs-2201-08821} on moltox21 and molbbbp. We follow the typical design for the predictor module as a $3$-layered MLP with ReLU activation in the intermediate layers. In our implementation, we set $\beta=\frac{2\times\hat{\beta}}{n\times(n-1)}$ (FIG-N) or $\beta=\frac{2\times\hat{\beta}}{r\times(r-1)}$ (FIG-VN). The $\alpha$ and $\hat{\beta}$ are searched between $[0.2, 2]$, step size $0.2$. In our implementation, the $K$ is set as $K=\texttt{round}(\hat{K}\times n)$ (for FIG-N) or $K=\texttt{round}(\hat{K}\times r)$ (for FIG-VN). $r$ is searched between $\{8, 16, 32\}$ for FIG-VN. We have a detailed sensitivity study to explore the best selection of $\hat{K}$ in Section~\ref{sec: sensitivity study}, which shows the best $\tilde{K}$ is around $0.75$.

\subsection{Implementation of Baseline Methods}
We search the number of layers of GIN~\cite{DBLP:conf/iclr/XuHLJ19}, GAT~\cite{DBLP:conf/iclr/VelickovicCCRLB18}, GATv2~\cite{DBLP:conf/iclr/Brody0Y22}, GatedGCN~\cite{DBLP:journals/corr/abs-1711-07553}, DIR~\cite{DBLP:conf/iclr/WuWZ0C22}, and GREA~\cite{DBLP:conf/kdd/LiuZXL022} between $\{2,3,5,10\}$ and report the best performance, considering configurations both with and without a virtual node connecting to all the given nodes.

Regarding the Transformer-based baselines (GT~\cite{DBLP:journals/corr/abs-2012-09699}, GraphiT~\cite{DBLP:journals/corr/abs-2106-05667}, SAN~\cite{DBLP:conf/nips/KreuzerBHLT21}, SAT~\cite{DBLP:conf/icml/ChenOB22}, Graphormer~\cite{DBLP:conf/nips/YingCLZKHSL21}, GraphTrans~\cite{DBLP:journals/corr/abs-2201-08821}, GPS~\cite{DBLP:conf/nips/RampasekGDLWB22}), for the (absolute or relative) positional encoding, we adhere to the suggestions made in their original papers. We also searched the number of layers between $\{2,3,5,10\}$.

Our GIN, GAT, GATv2, and GatedGCN implementations are from the PyTorch-geometric package. Our implementations of GT~\footnote{\url{https://github.com/graphdeeplearning/graphtransformer}}, GraphiT~\footnote{\url{https://github.com/inria-thoth/GraphiT}}, SAN~\footnote{\url{https://github.com/DevinKreuzer/SAN}}, SAT~\footnote{\url{https://github.com/BorgwardtLab/SAT}}, Graphormer~\footnote{\url{https://github.com/microsoft/Graphormer}}, GraphTrans~\footnote{\url{https://github.com/ucbrise/graphtrans}}, GPS~\footnote{\url{https://github.com/rampasek/GraphGPS}}, DIR~\footnote{\url{https://github.com/Wuyxin/DIR-GNN}}, GREA~\footnote{\url{https://github.com/liugangcode/GREA}} are adapted from publicly available code.

\section{Algorithms}
A step-by-step algorithm for FIG-VN is provided in Algorithm~\ref{alg: FIG-VN training}. 

\begin{algorithm2e}[!t]
\caption{FIG-VN single training step for every training graph $\mathcal{G}$}
\label{alg: FIG-VN training}
\SetAlgoLined
\SetKwInOut{Input}{Input}\SetKwInOut{Output}{Output}
\Input{a labelled graph $(\mathcal{G}, y)$, a sampled graph $\tilde{\mathcal{G}}$ from the same batch as $\mathcal{G}$, $\theta=\{\theta_{\texttt{enc}}, \theta_{\texttt{aug-VN}},\theta_{\texttt{pred}}\}$, $\phi$;}
\Output{updated $\theta$ and $\phi$;}

compute the node embedding matrices $\mathbf{H}=\theta_{\texttt{enc}}(G)$ and $\tilde{\mathbf{H}}=\theta_{\texttt{enc}}(\tilde{G})$\;
compute rationale and environment embeddings as $(\mathbf{H}_{\texttt{ra}}, \mathbf{H}_{\texttt{env}})=\theta_{\texttt{aug-VN}}(\mathbf{H})$, $(\tilde{\mathbf{H}}_{\texttt{ra}}, \tilde{\mathbf{H}}_{\texttt{env}})=\theta_{\texttt{aug-VN}}(\tilde{\mathbf{H}})$ via Eqs.~\eqref{eq: virtual node embeddings},~\eqref{eq: rationale virtual node embeddings FIG-VN}, and~\eqref{eq: environment virtual node embeddings FIG-VN}\;
concatenate rationale-environment pairs $\mathbf{H}_{\texttt{ra}}||\mathbf{H}_{\texttt{env}}$ and $\mathbf{H}_{\texttt{ra}}||\tilde{\mathbf{H}}_{\texttt{env}}$\;
compute $\mathcal{L}_{\texttt{util}}$ via Eq.~\eqref{eq: FIG-N utility loss}\;
compute $\mathcal{L}_{\texttt{reg}}$ via Eqs.~\eqref{eq: node-level regularization} and~\eqref{eq: FIG-N indicator vector}\;
% compute gradients $\frac{\partial (\mathcal{L}_{\texttt{util}}+\beta\mathcal{L}_{\texttt{reg}})}{\partial \theta}$ and $\frac{\partial (\mathcal{L}_{\texttt{util}}+\beta\mathcal{L}_{\texttt{reg}})}{\partial \phi}$\;
update $\theta$ via gradient descent with $\frac{\partial (\mathcal{L}_{\texttt{util}}+\beta\mathcal{L}_{\texttt{reg}})}{\partial \theta}$\;
update $\phi$ via gradient ascent with $\frac{\partial (\mathcal{L}_{\texttt{util}}+\beta\mathcal{L}_{\texttt{reg}})}{\partial \phi}$\;
\end{algorithm2e}

\section{Additional Experiments}

\subsection{Sensitivity Study}
\label{sec: sensitivity study}
\begin{figure}[t!]
\centering
\begin{subfigure}{.24\columnwidth}
  \centering
  \includegraphics[width=\columnwidth]{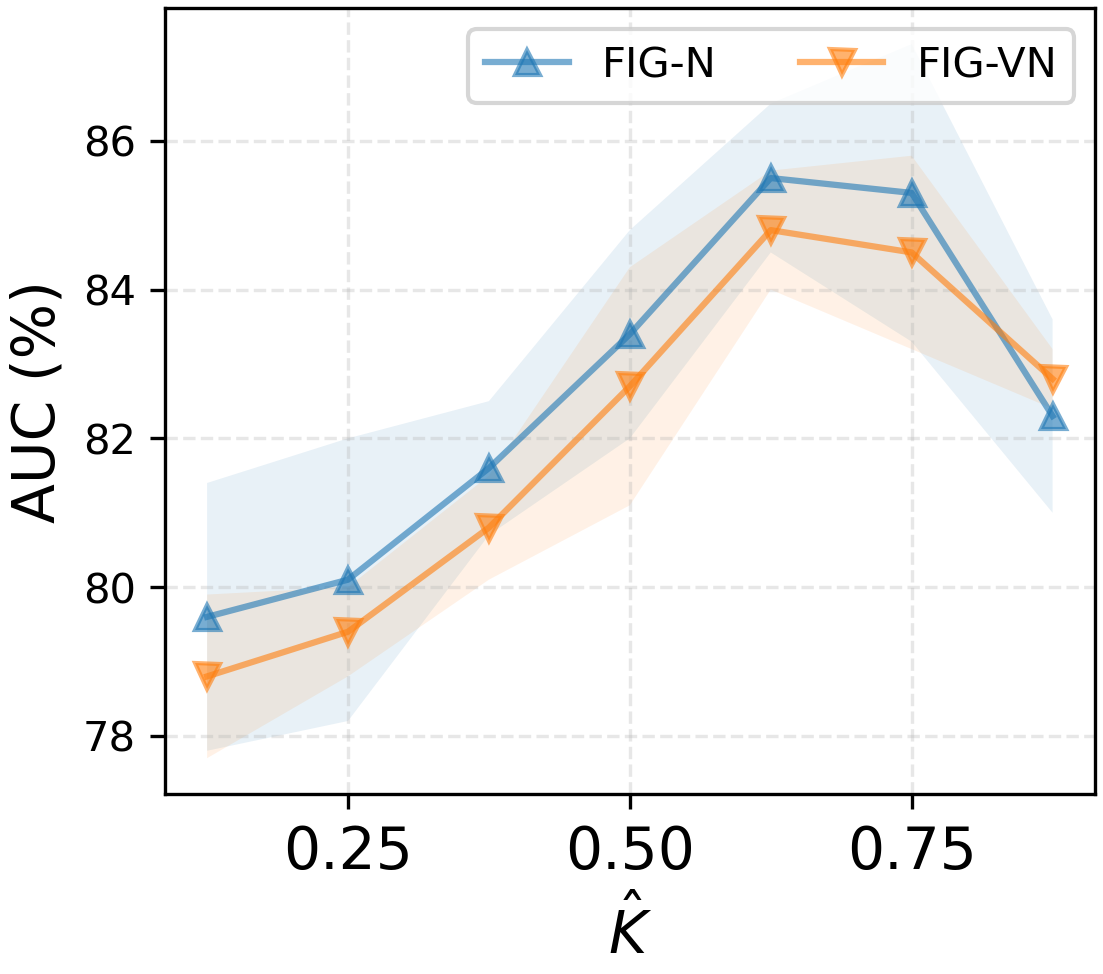}
  \caption{molbace}
\end{subfigure}
% \hfill
\begin{subfigure}{.24\columnwidth}
  \centering
  \includegraphics[width=\columnwidth]{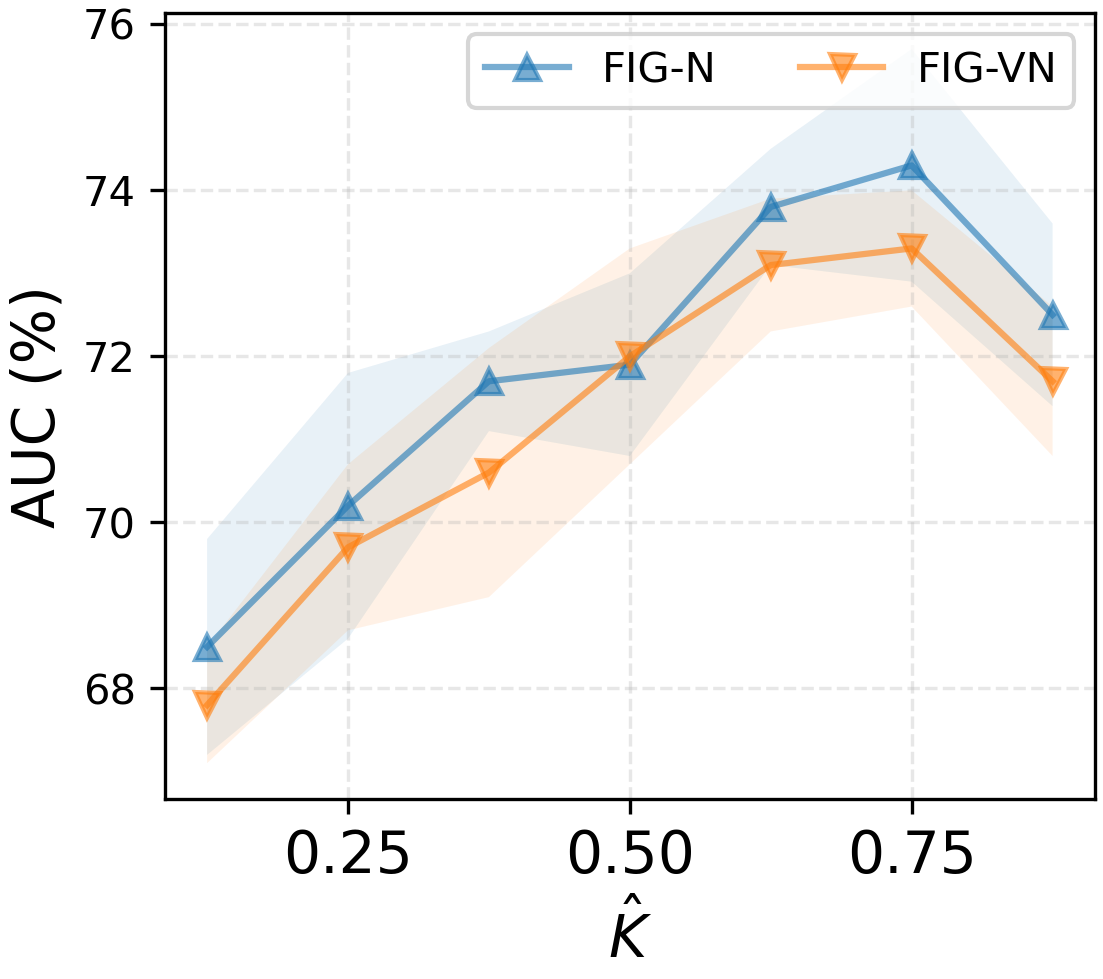}
  \caption{molbbbp}
\end{subfigure}
\caption{Performance of FIG-N/VN with different $\hat{K}$.}
\label{fig: sensitivity study}
\end{figure}

In this section, we carefully study the impact of hyperparameter $K$ (from Eq.~\eqref{eq: rationale node embeddings},~\eqref{eq: environment node embeddings},~\eqref{eq: rationale virtual node embeddings FIG-VN}, and~\eqref{eq: environment virtual node embeddings FIG-VN}), which determines the ratio of the rationale and environment subgraphs. In our implementation, we set $K=\texttt{round}(\hat{K}\times n)$ (for FIG-N) or $K=\texttt{round}(\hat{K}\times r)$ (for FIG-VN). We evaluate the model performance across varying $\hat{K}$ on the molbace and molbbbp datasets in Figure~\ref{fig: sensitivity study}. We note that the model performance degrades if most nodes/virtual nodes are marked as the environment component. Similar performance degradation is observed if too many nodes/virtual nodes are marked as the rationale nodes (e.g., $\hat{K}=0.875$). That is because for a large $\hat{K}$ (e.g., $\hat{K}=1$), the model degenerates to a vanilla graph encoder, with less intervention involved. The best performance is observed when $\hat{K}$ is set as $0.75$ or $0.675$.

\subsection{Training Convergence}
\label{sec: training convergence}
\begin{figure*}[t!]
\centering
\begin{subfigure}{.33\textwidth}
  \centering
  \includegraphics[width=1\linewidth]{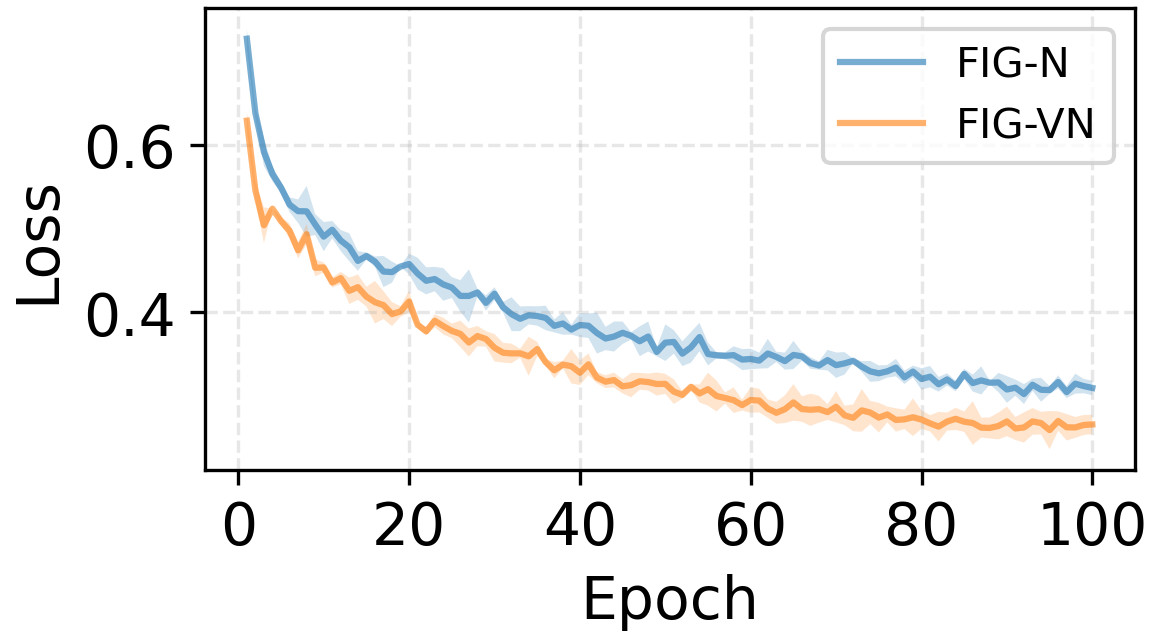}
  \caption{molbace, GIN}
\end{subfigure}
\hfill
\begin{subfigure}{.33\textwidth}
  \centering
  \includegraphics[width=1\linewidth]{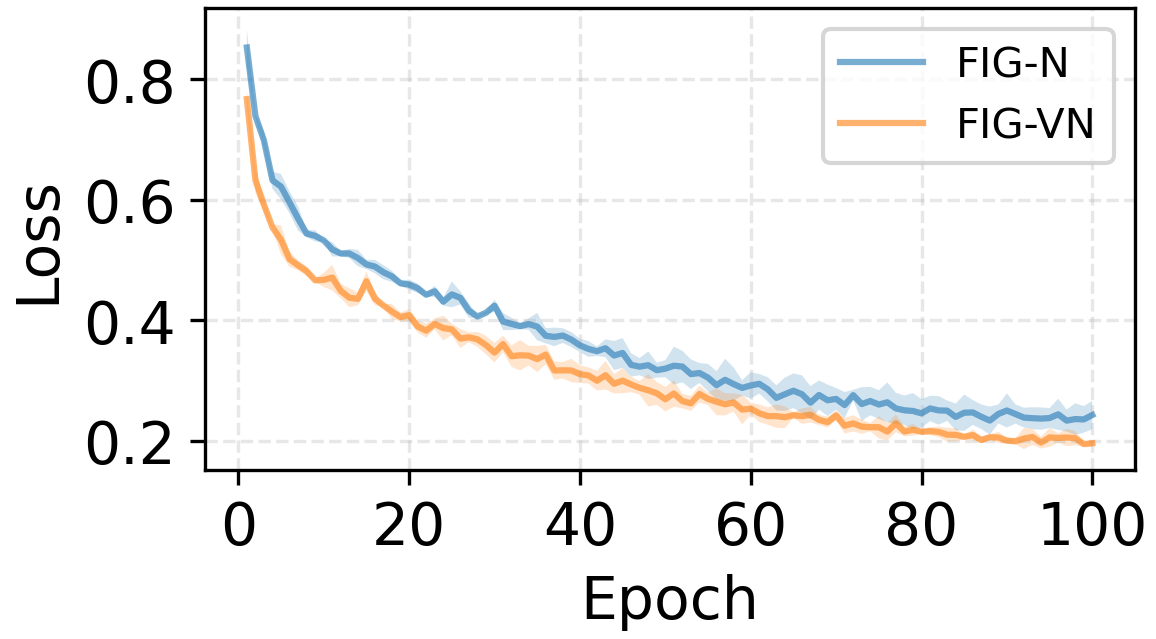}
  \caption{molbace, GPS}
\end{subfigure}
\begin{subfigure}{.33\textwidth}
  \centering
  \includegraphics[width=1\linewidth]{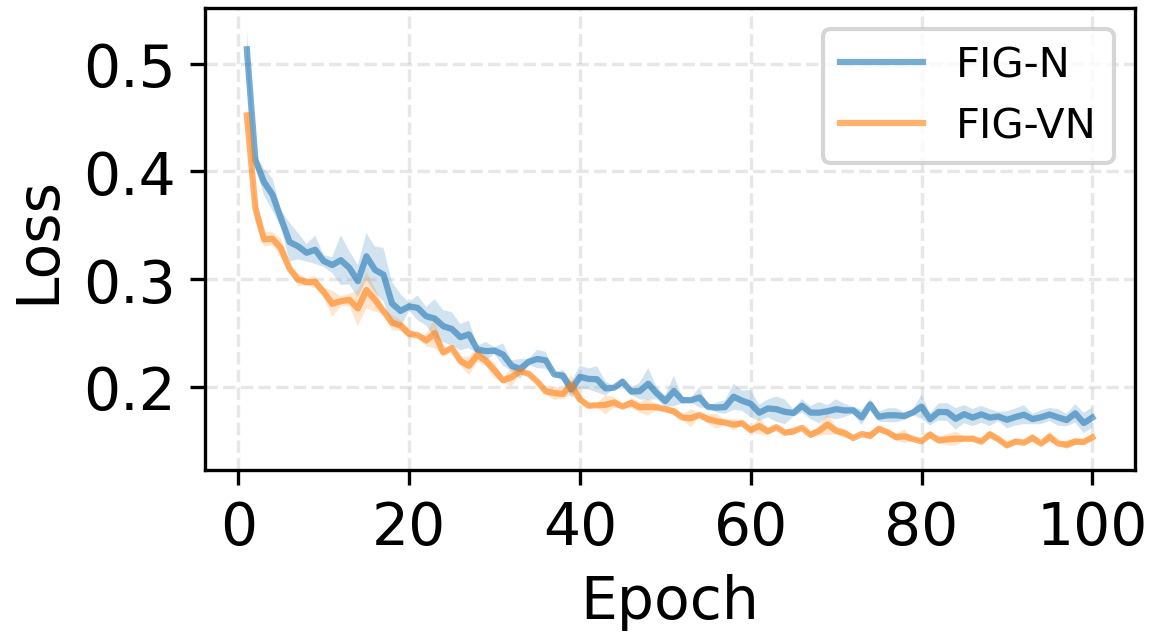}
  \caption{molbbbp, GIN}
\end{subfigure}
\hfill
\begin{subfigure}{.33\textwidth}
  \centering
  \includegraphics[width=1\linewidth]{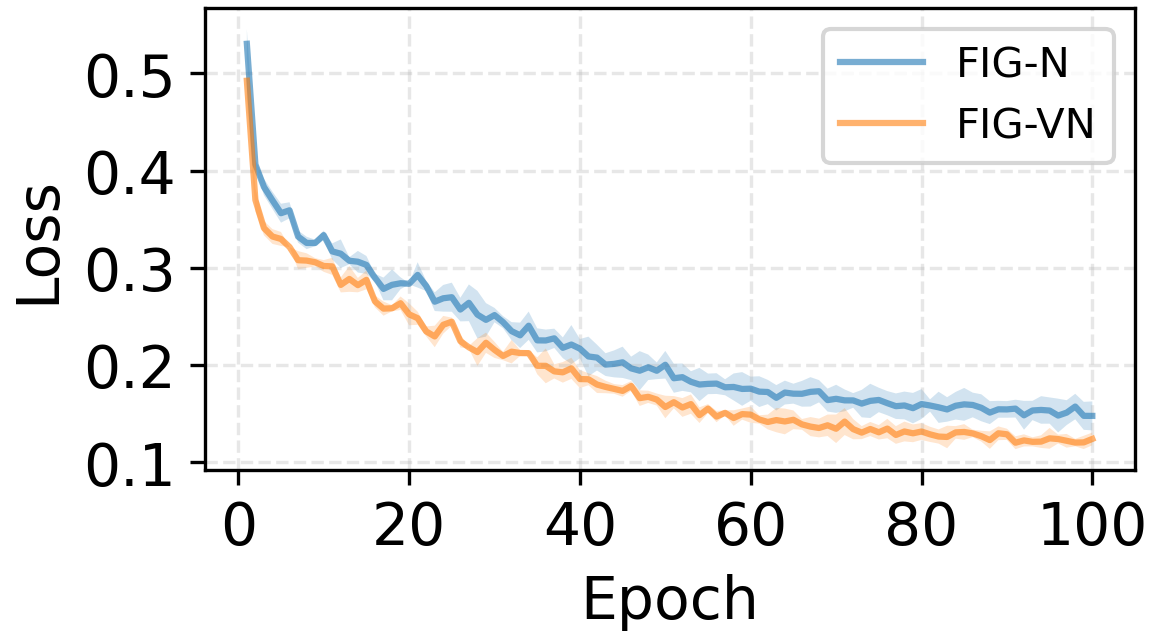}
  \caption{molbbbp, GPS}
\end{subfigure}
\begin{subfigure}{.33\textwidth}
  \centering
  \includegraphics[width=1\linewidth]{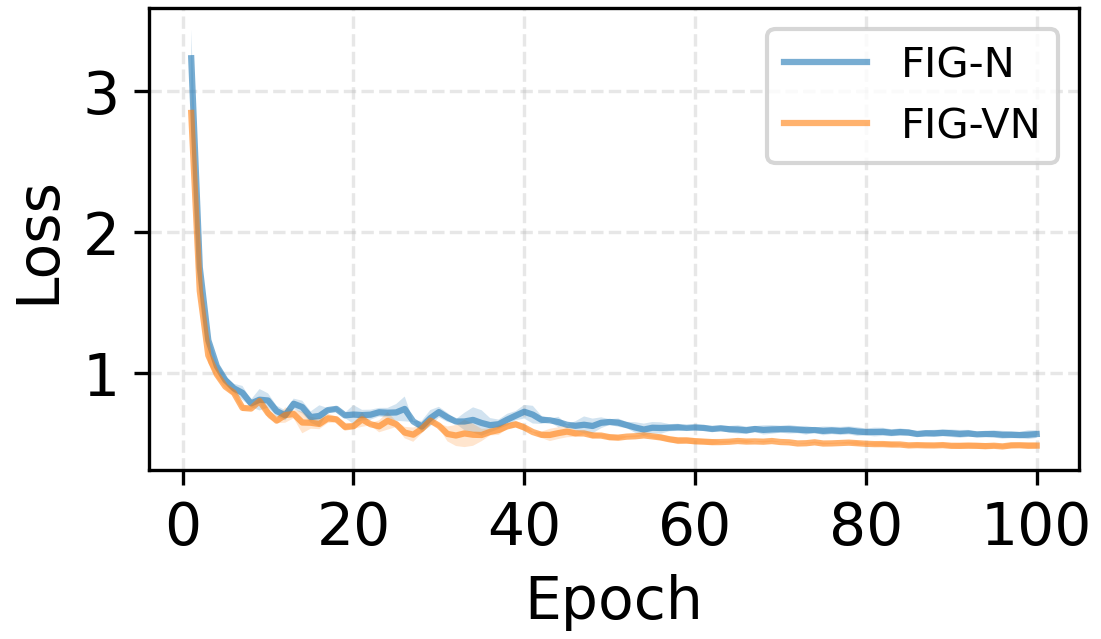}
  \caption{mollipo, GIN}
\end{subfigure}
\hfill
\begin{subfigure}{.33\textwidth}
  \centering
  \includegraphics[width=1\linewidth]{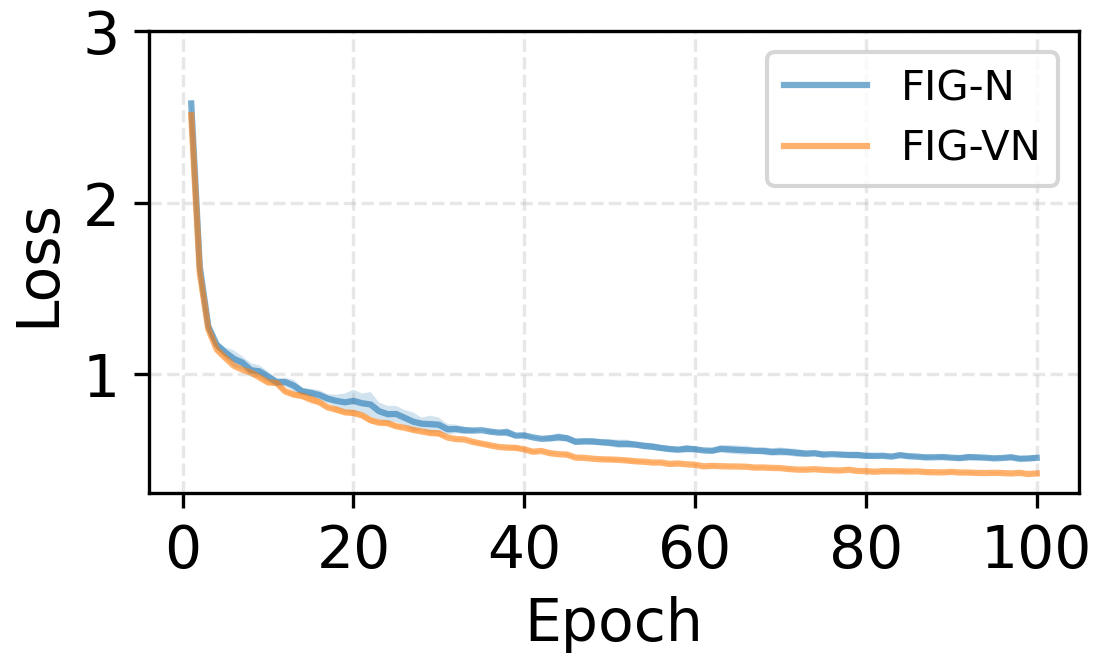}
  \caption{mollipo, GPS}
\end{subfigure}
\caption{Training loss of FIG-N/VN with different datasets and encoders.}
\label{fig: training convergence}
\end{figure*}

We are concerned about the impact of the min-max objective on the training stability of FIG-N and FIG-VN. We monitor the training losses of both FIG-N and FIG-VN across three datasets (molbace, molbbbp, mollipo) using two encoders (GIN and GPS). The results, presented in Figure~\ref{fig: training convergence}, demonstrate that the training remains stable even when $\theta$ (representing the encoder, augmenter, and predictor) and $\phi$ (representing the intervener) engage in a min-max game.

\subsection{Attention Visualization} 
\begin{figure}[t!]
	\centering
	\includegraphics[width=.6\columnwidth]{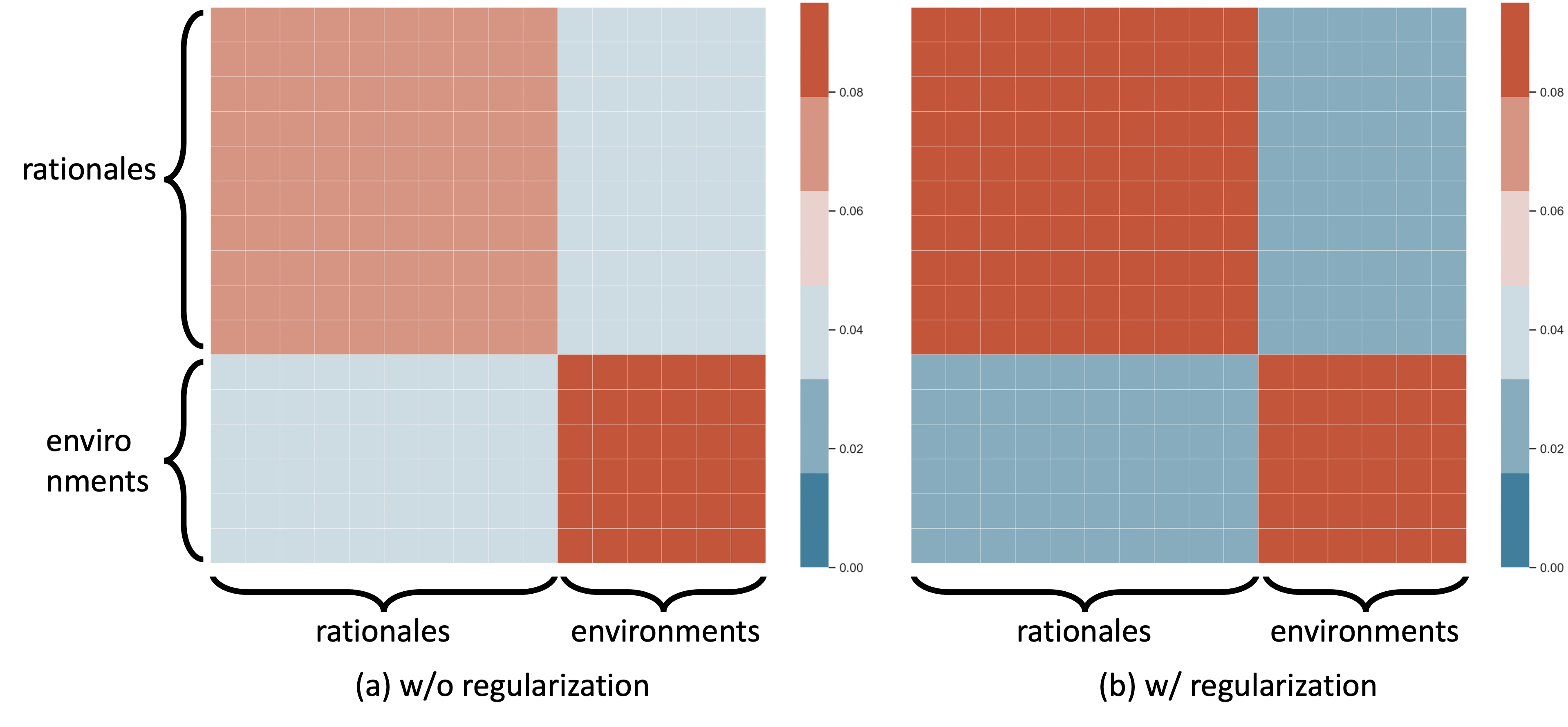}
	\caption{Heatmap of the adjacency matrix of the intervener $\phi$. (a) without the regularization term Eq.~\eqref{eq: regularization term} and (b) with the regularization term Eq.~\eqref{eq: regularization term}.}
	\label{fig: heatmap}
\end{figure}

In this section, we aim to evaluate the significance of the regularization term by visualizing the adjacency matrix $\mathbf{P}$, effectively the attention matrix, of the intervener $\phi$ in Figure~\ref{fig: heatmap}. For clarity in visualization, we choose FIG-VN. Unlike FIG-N, which works on a variable number of nodes (from different graphs), FIG-VN maps nodes to a predetermined number of virtual nodes, simplifying the presentation. Specifically, we set the number of virtual nodes $r$ to $16$ with $K$ at $10$, designating $10$ virtual nodes to rationales and the remainder as environments. All the visualization results are obtained from the molbace dataset. It is worth noting that the attention matrix $\mathbf{P}$ is normalized row-wise by $\mathtt{Softmax}$.
From our observations, we highlight two primary insights:
\begin{itemize}
    \item Interestingly, even in the absence of the regularization term, in Figure~\ref{fig: heatmap}(a), interactions between rationales and environments appear significantly weaker than those within the rationales themselves. One potential explanation is the changing nature of the environment. In optimizing the utility loss $\mathcal{L}_{\texttt{util}} = \mathcal{L}_{\texttt{task}}(\mathbf{H}_{\texttt{ra}}||\mathbf{H}_{\texttt{env}})+\alpha\mathcal{L}_{\texttt{task}}(\mathbf{H}_{\texttt{ra}}||\tilde{\mathbf{H}}_{\texttt{env}})$, the ever-changing environment ($\tilde{\mathbf{H}}_{\texttt{env}}$) might lead the model to minimize interactions between rationales and environments so that the utility of the rationale can be ensured.
    \item The first observation supports our decision to introduce the regularization term, which aims to penalize rationale-environment interactions. When the proposed regularization term (Eq.~\eqref{eq: regularization term}) is implemented, in Figure~\ref{fig: heatmap}(b), there is a noticeable decline in rationale-environment interactions (the off-diagonal blocks in Figure~\ref{fig: heatmap}). As demonstrated in our earlier ablation study, this leads to enhanced model performance.
\end{itemize}

\subsection{Rationale Visualization}

\begin{figure}[t!]
\centering
\begin{subfigure}{.25\columnwidth}
  \centering
  \includegraphics[width=1\linewidth]{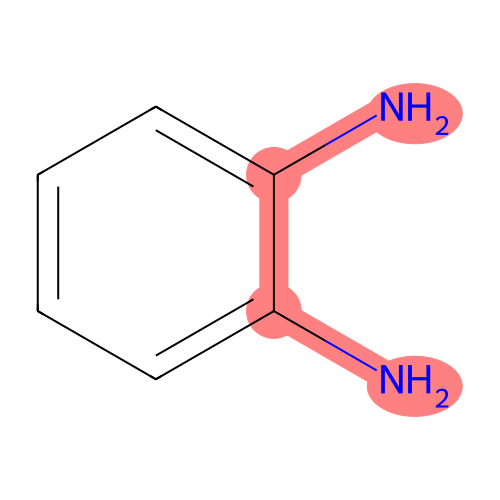}
\end{subfigure}
\begin{subfigure}{.25\columnwidth}
  \centering
  \includegraphics[width=1\linewidth]{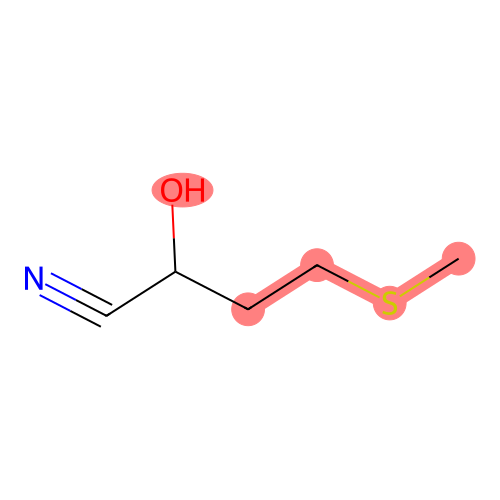}
\end{subfigure}
\begin{subfigure}{.25\columnwidth}
  \centering
  \includegraphics[width=1\linewidth]{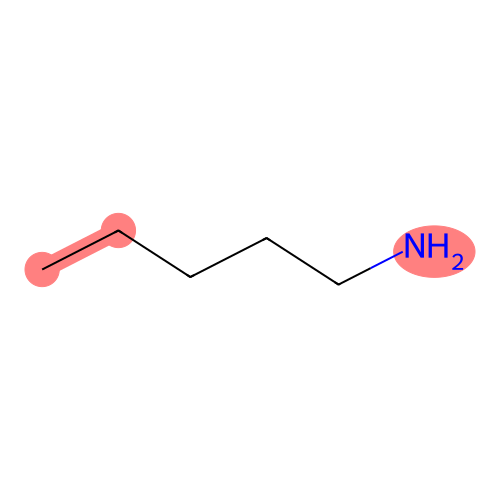}
\end{subfigure}
\hfill
\begin{subfigure}{.25\columnwidth}
  \centering
  \includegraphics[width=1\linewidth]{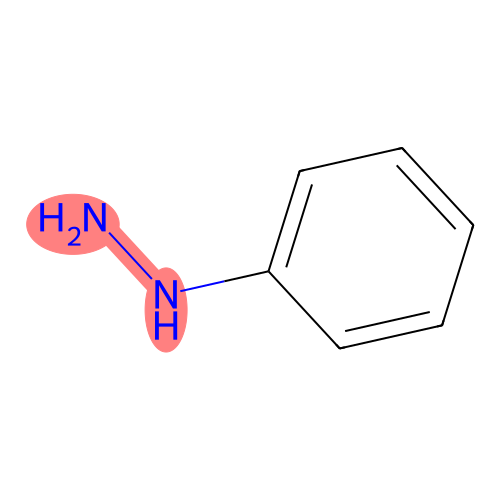}
\end{subfigure}
\begin{subfigure}{.25\columnwidth}
  \centering
  \includegraphics[width=1\linewidth]{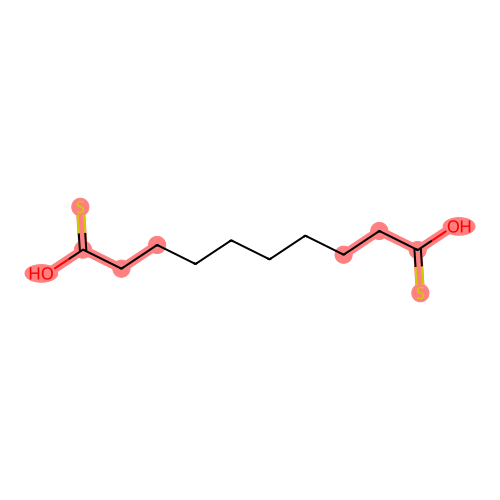}
\end{subfigure}
\begin{subfigure}{.25\columnwidth}
  \centering
  \includegraphics[width=1\linewidth]{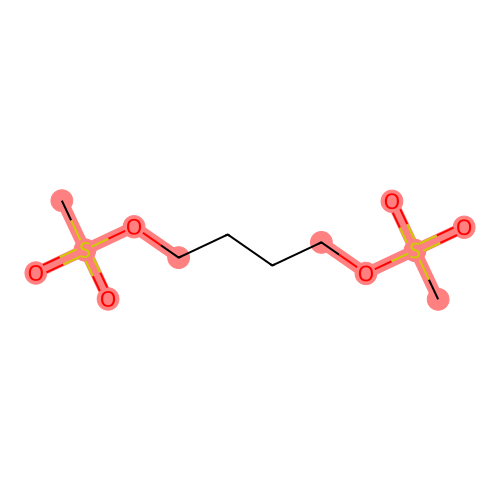}
\end{subfigure}
\caption{Rationales highlighted by FIG-N.}
\label{fig: rationale vis}
\end{figure}

Figure~\ref{fig: rationale vis} presents examples showing the rationale detected by our FIG-N model on the AQSOL dataset~\cite{DBLP:journals/jmlr/DwivediJL0BB23}.

\section{Soft argtop-$K$ Trick}
\label{sec: argtopK}

In the main content, for the FIG-N model, the augmenter will generate the partition vector $\mathbf m$ which is then used to partition the node embedding matrix via selecting the top-$K$ indices from $\mathbf m$:
\begin{align}
    \mathbf{m}& = \mathtt{Sigmoid}(\mathtt{MLP}(\mathbf{H}, \theta_{\texttt{aug-N}}))\in\mathbb R^n\\
    \texttt{idx}_{\texttt{ra}} &= \mathtt{argtopK}(\mathbf{m})\in\mathbb N_+^K\\
    \mathbf{H}_{\texttt{ra}}&=\mathbf{H}[\texttt{idx}_\texttt{ra}, :]\in\mathbb{R}^{K\times d}\\
    \mathbf{H}_{\texttt{env}}&=\mathbf{H}[\texttt{idx}_\texttt{env}, :]\in\mathbb{R}^{(n-K)\times d}
\end{align}
The hard $\mathtt{argtopK}$ breaks the differentiability of the model so that $\theta_{\texttt{aug-N}}$ has no gradient. Here, we present a soft argtop-$K$ trick which is inspired by the soft top-K trick\footnote{\url{https://github.com/ZIB-IOL/merlin-arthur-classifiers/blob/main/soft-topk.py}}. Overall, the key ideas are as follows,
\begin{enumerate}
    \item The index vector $\texttt{idx}_{\texttt{ra}}\in\mathbb N_+^K$ can be viewed as a list of one-hot index encoding $\mathbf {idx}\in\{0,1\}^{K\times n}$ whose every row is a one-hot vector. If the $i$-th row's $j$-th element is $1$, it means the $j$-th element in $\mathbf m$ is the $i$-th largest element in $\mathbf m$. Then, we can use the matrix multiplication to index the matrix $\mathbf H$:
    \begin{align}
        \mathbf{H}[\texttt{idx}_\texttt{ra}, :] \Longleftrightarrow \mathbf{idx}\times \mathbf H \in\mathbb R^{K\times d}
    \end{align}
    \item We aim to find a soft and differentiable matrix to approximate $\mathbf {idx}$. The trick is to use the fact that every row of $\mathbf {idx}$ is one-hot, which can be approximated by the output of the softmax function. Hence, the implementation is to call the softmax $K$ times repeatedly.
    
    \item The parameterization trick\footnote{\url{https://pytorch.org/docs/stable/generated/torch.nn.functional.gumbel_softmax.html}} can be used: y\_hard - y\_soft.detach() + y\_soft, whose main idea is to ensure (1) the forward process to use the hard indexing and (2) the backpropagation to update the soft indices.
\end{enumerate}

We provide the exemplar code as follows.
\begin{python}
import torch

def soft_arg_top_K(K,H,m):
    """
    Select the argtop K indices from the vector m, and index the node embedding matrix H.
    
    K: number of nodes selected
    H: node embedding matrix (n, d), n: number of nodes, d: hidden dimension
    m: masking vector (n,)
    """
    
    top_K_indices = []
    for i in range(K):
        top1_index_soft = torch.nn.Softmax(m)
        top1_index_hard = torch.zeros_like(top1_index_soft)
        top1_index_hard[torch.argmax(top1_index_soft)] = 1
        top1_index = top1_index_hard - top1_index_soft.detach() + top1_index_soft
        top_K_indices.append(top1_index)
        m = m - top1_index_hard * 1e6  # mask the selected entry
    top_K_indices = torch.stack(top_K_indices, dim=0)
    H_ra = torch.mm(top_K_indices, H)
    return H_ra
\end{python}

\end{document}